\pdfoutput=1

\documentclass[11pt]{article}

\usepackage[final]{acl}
\usepackage{arydshln}
\usepackage{times}
\usepackage{latexsym}

\usepackage[T1]{fontenc}

\usepackage[utf8]{inputenc}

\usepackage{microtype}

\usepackage{inconsolata}

\usepackage{tikz}
\usepackage{graphicx}
\usepackage{multirow}
\usepackage{amsmath}
\usepackage{amssymb}
\usepackage{wrapfig}   
\usepackage{float}
\usepackage{algorithm}
\usepackage{algorithmic}
\usepackage{mdframed}
\usepackage{tcolorbox}
\usepackage{subcaption}
\usepackage{listings}
\usepackage{lipsum}
\usepackage{array}
\usepackage{chngcntr}
\usepackage{ragged2e}
\tcbuselibrary{breakable}
\usepackage{booktabs}
\usepackage{makecell}
\usepackage{caption}
\usepackage{xcolor} 
\usepackage{pifont}  

\usepackage[table]{xcolor} 
\definecolor{lightgray}{gray}{0.9} 
\usepackage{tcolorbox}
\tcbuselibrary{skins}

\usepackage{xstring}   

\definecolor{myGreen}{RGB}{0, 150, 0}
\definecolor{myRed}{RGB}{200, 0, 0}
\newcommand{\perf}[2]{%
    #1%
    \rlap{$
        \,_{\IfBeginWith{#2}{-}%
            {\color{myGreen}\text{\tiny{(#2)}}}%
            {\color{myRed}\text{\tiny{(#2)}}}%
        }
    $}%
}

\newtcolorbox{definitionbox}[1][]{%
  colback=blue!5,       
  colframe=blue!50!black, 
  coltitle=white,       
  colbacktitle=blue!60!black, 
  boxrule=1.5pt,                 
  rounded corners,               
  fonttitle=\bfseries,  
  enhanced,
  attach boxed title to top left={yshift=-2mm,xshift=2mm},
  boxed title style={
    rounded corners,
    borderline west={0pt}{0pt}{white}, 
    borderline east={0pt}{0pt}{white},
    borderline north={0pt}{0pt}{white},
    borderline south={0pt}{0pt}{white},
  },
  title=Definition,
  #1
}

\title{KnowRL: Exploring Knowledgeable Reinforcement Learning for Factuality}

\author{
    Baochang Ren\textsuperscript{$\spadesuit\heartsuit$}, 
    Shuofei Qiao\textsuperscript{$\spadesuit\heartsuit$}, 
    Ningyu Zhang\textsuperscript{$\spadesuit\heartsuit$}, 
    \textbf{Da Zheng}\textsuperscript{$\diamondsuit\heartsuit$}, 
    \textbf{Huajun Chen}\textsuperscript{$\spadesuit\heartsuit$}\thanks{Corresponding author.}\\
    \textsuperscript{$\spadesuit$} Zhejiang University \quad
    \textsuperscript{$\diamondsuit$} Ant Group \quad \\
     \textsuperscript{$\heartsuit$} Zhejiang University - Ant Group Joint Laboratory of Knowledge Graph\\
    \texttt{\{baochang.ren, shuofei, zhangningyu\}@zju.edu.cn}
}

\begin{document}
\maketitle

\begin{abstract} 
Slow-thinking Large Language Models (LLMs) have demonstrated strong reasoning capabilities but often suffer from severe hallucinations due to an inability to recognize their knowledge boundaries. Existing Reinforcement Learning (RL) approaches typically rely on outcome-oriented rewards, which can inadvertently reinforce fabricated reasoning paths when the final answer is correct. To address this, we propose \textbf{Know}ledge-enhanced \textbf{RL}, \textbf{KnowRL}, a framework that integrates factual supervision directly into the reasoning process. By decomposing the chain of thought into atomic facts and verifying them against the corresponding ground-truth knowledge, KnowRL performs fine-grained checks to encourage models to reason faithfully. Crucially, this process-oriented supervision teaches the model to identify its knowledge boundaries, learning to say ``I don't know'' instead of fabricating answers when information is missing. Experimental results demonstrate that KnowRL effectively mitigates hallucinations---reducing the Incorrect Rate on SimpleQA by 20.3\% for distillation-based slow-thinking models while maintaining strong performance on complex reasoning benchmarks like GPQA and AIME 2025. Furthermore, our method shows robust transferability to out-of-distribution tasks, indicating that the model learns a generalizable verification behavior\footnote{\url{https://github.com/zjunlp/KnowRL}.}. 
\end{abstract}

\section{Introduction}
\begin{figure}[t]
    \centering
    \resizebox{.45\textwidth}{!}{
    \includegraphics{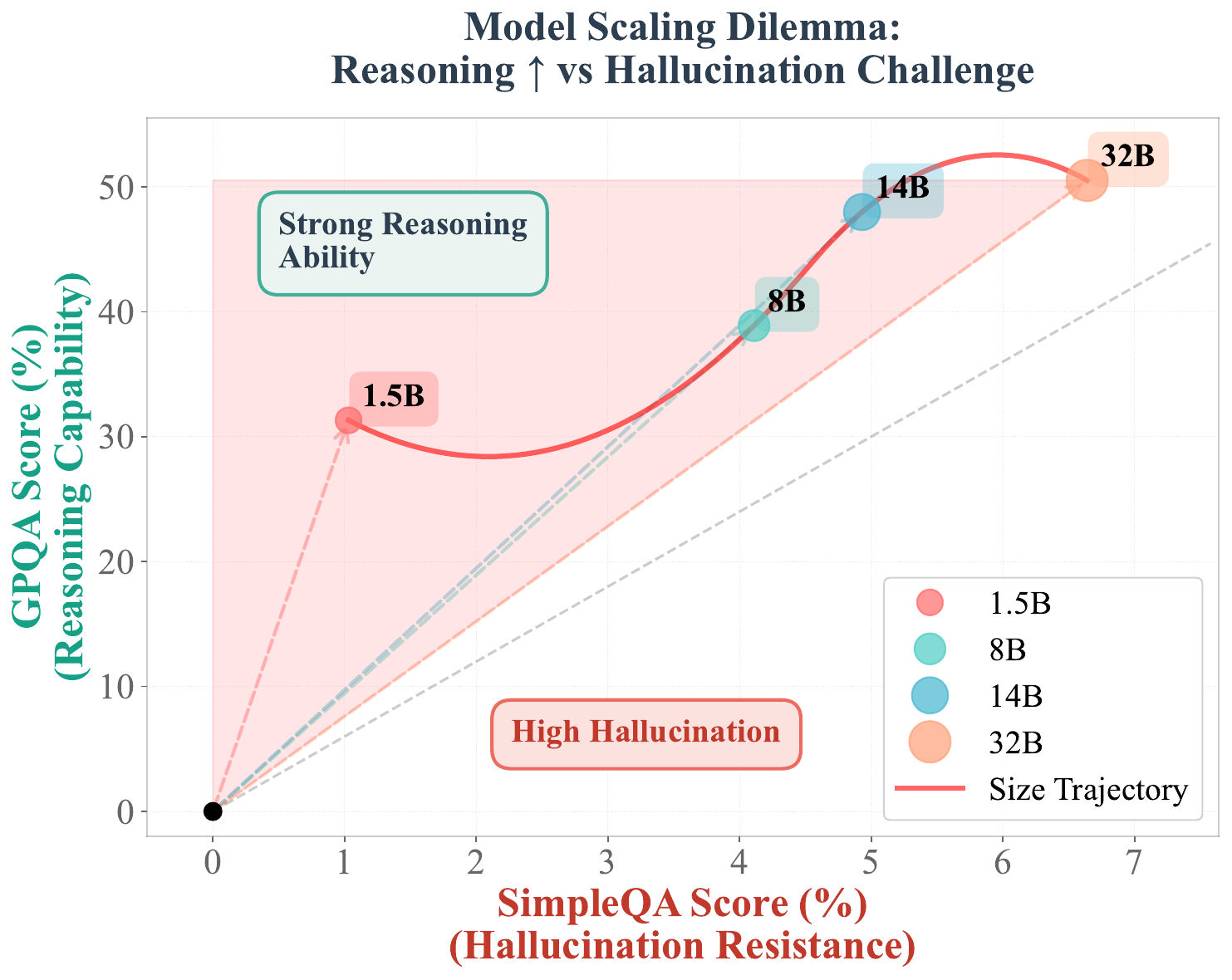}}
    \caption{Scaling improves reasoning ability (GPQA) but does not reduce hallucinations (SimpleQA). Results are shown for DeepSeek-R1-Distill models of varying sizes (Qwen-1.5B, Llama-8B, Qwen-14B, Qwen-32B).}
    \vspace{-3ex}
    \label{fig:scaling}
\end{figure}

Recent advancements, represented by models like DeepSeek-R1 \citep{guo2025deepseek}, mark a paradigm shift towards ``slow thinking''. 
By leveraging Reinforcement Learning (RL) to encourage extended Chains of Thought (CoT), these models have achieved remarkable breakthroughs in complex reasoning tasks. 
However, a critical paradox has emerged: while scaling reasoning compute significantly boosts problem-solving abilities, it does not naturally align with factual reliability. 
In fact, slow thinking models often exhibit severe hallucinations \citep{heyman2025reasoning,patel2024multi,arcuschin2025chain}. 
As shown in Figure~\ref{fig:scaling}, larger reasoning models achieve higher GPQA~\citep{rein2024gpqa} scores but fail to improve or even regress on hallucination benchmarks like SimpleQA \citep{wei2024measuring}. 
For instance, the DeepSeek-R1-Distill-Qwen-32B \citep{guo2025deepseek} achieves an accuracy of only 6.64\% on the SimpleQA dataset. 
This suggests that without proper guidance, the long reasoning process can turn into a "snowball" of errors, where one small mistake leads to a completely fabricated conclusion. 
This raises a critical question: \textit{\textbf{Why do models with such strong reasoning abilities still fail so badly at factual reliability?}}


\begin{figure}[t!]
    \centering
    \resizebox{.45\textwidth}{!}{
    \includegraphics{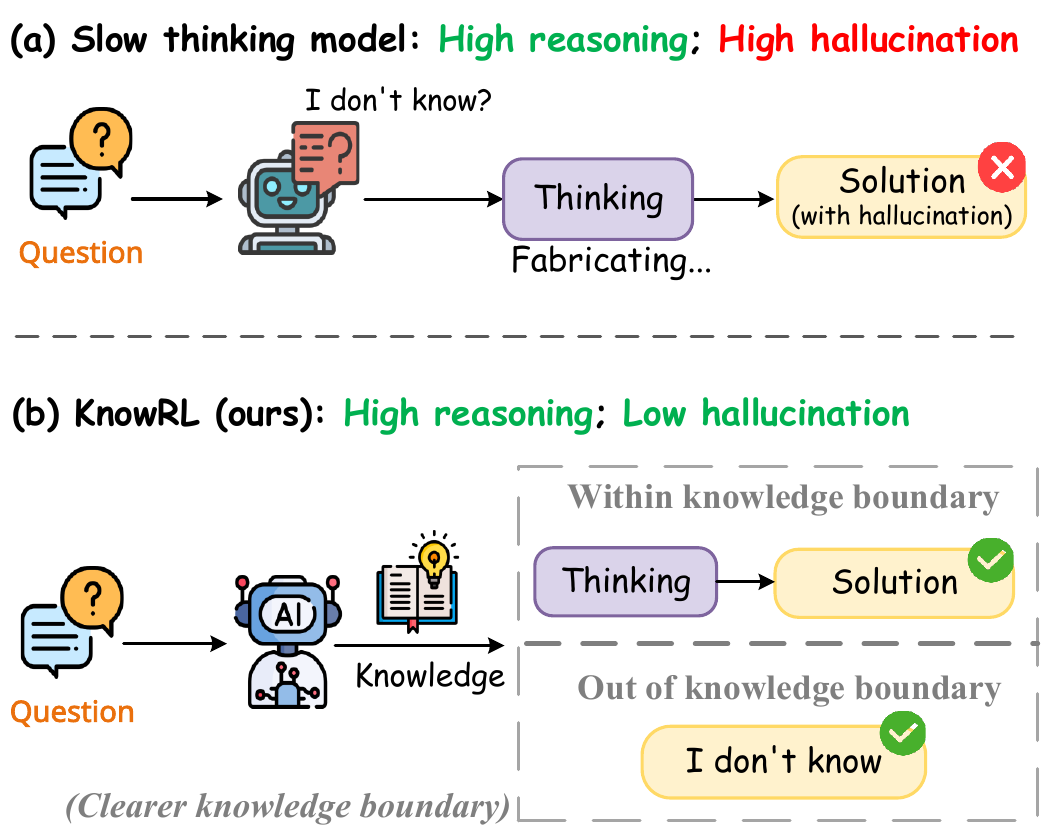}}
    \caption{KnowRL reduces hallucinations in slow-thinking models.}
     \vspace{-3ex}
    \label{fig:intro}
\end{figure}

The root cause lies in the way we currently train these models.
Standard RL heavily relies on outcome-oriented rewards, which optimize for the final answer while treating the reasoning process as a black box. 
This approach has two main flaws.
\textit{First}, it ignores the Knowledge Boundary---the model's ability to distinguish between what it knows and what it does not know. 
Without this boundary, models try to guess the answer to get the reward. 
\textit{Second}, it creates a supervision gap: model might generate a correct answer using wrong or made up reasoning steps. 
Since the reward signal depends solely on the final outcome, the model mistakenly learns that fabricating reasoning paths is a valid strategy; therefore, the RL algorithm unwittingly reinforces this hallucinated logic.
Consequently, the model becomes ``smart'' at reasoning but ``dishonest'' about facts.

Addressing this challenge with existing methods proves difficult. 
Retrieval Augmented Generation \citep[RAG;][]{lewis2020retrieval} faces significant retrieval efficiency bottlenecks when integrated into the extensive reasoning steps of slow thinking models. 
Supervised Fine Tuning \citep[SFT;][]{ouyang2022training} primarily relies on static knowledge injection. 
However, this paradigm suffers from severe catastrophic forgetting \citep{chen2025retaining}, often degrading the model's inherent reasoning capabilities. 
More fundamentally, SFT encourages the model to merely memorize static knowledge rather than learning the generalized behavior of reasoning.
Consequently, it fails to instill the critical strategy of ``knowing what you know and what you do not know''---a dynamic judgment of Knowledge Boundaries. 
In contrast, Reinforcement Learning is uniquely suited to shape this behavioral strategy \citep{gandhi2025cognitive}.
Therefore, there is an urgent need for a RL framework that can inherently instill factual discipline into the model's reasoning behaviors without compromising its reasoning capabilities.

So, to bridge this gap, we propose Knowledge enhanced Reinforcement Learning (KnowRL), a framework that integrates factuality supervision directly into the RL reasoning loop. 
Unlike outcome-based approaches, KnowRL opens the ``black box'' of thinking. 
By decomposing the Chain of Thought into atomic facts and verifying them against the corresponding ground-truth knowledge, KnowRL provides dense, process level rewards. 
This design transforms the RL objective: instead of merely ``getting the answer right,'' the model is incentivized to reason faithfully and, crucially, to recognize its knowledge boundaries---learning to say ``I don't know'' rather than fabricating a plausible sounding response when information is missing, as illustrated in Figure~\ref{fig:intro}.

Experimental results validate the effectiveness of KnowRL. 
On hallucination benchmarks, KnowRL significantly reduces the error rate. 
For instance, dropping the Incorrect Rate on SimpleQA by 20.3\% for 7B slow thinking model. 
Crucially, this gain in factuality does not come at the cost of reasoning ability; KnowRL maintains strong performance on complex reasoning datasets like GPQA and AIME 2025. 
Furthermore, our method demonstrates strong transferability and Out-Of-Distribution (OOD) generalization to knowledge domains outside the training distribution; it significantly improves performance on ChineseSimpleQA~\citep{he2024chinese} even when the primary knowledge source is purely English-based. 
These findings suggest that KnowRL helps the model internalize a universal verification behavior rather than merely memorizing language-specific facts.  
These results confirm that KnowRL successfully aligns the slow thinking process with factual accuracy, offering a robust path for building reliable reasoning models.

\begin{figure*}[t!]
    \centering
    \includegraphics[width=1.0\linewidth]{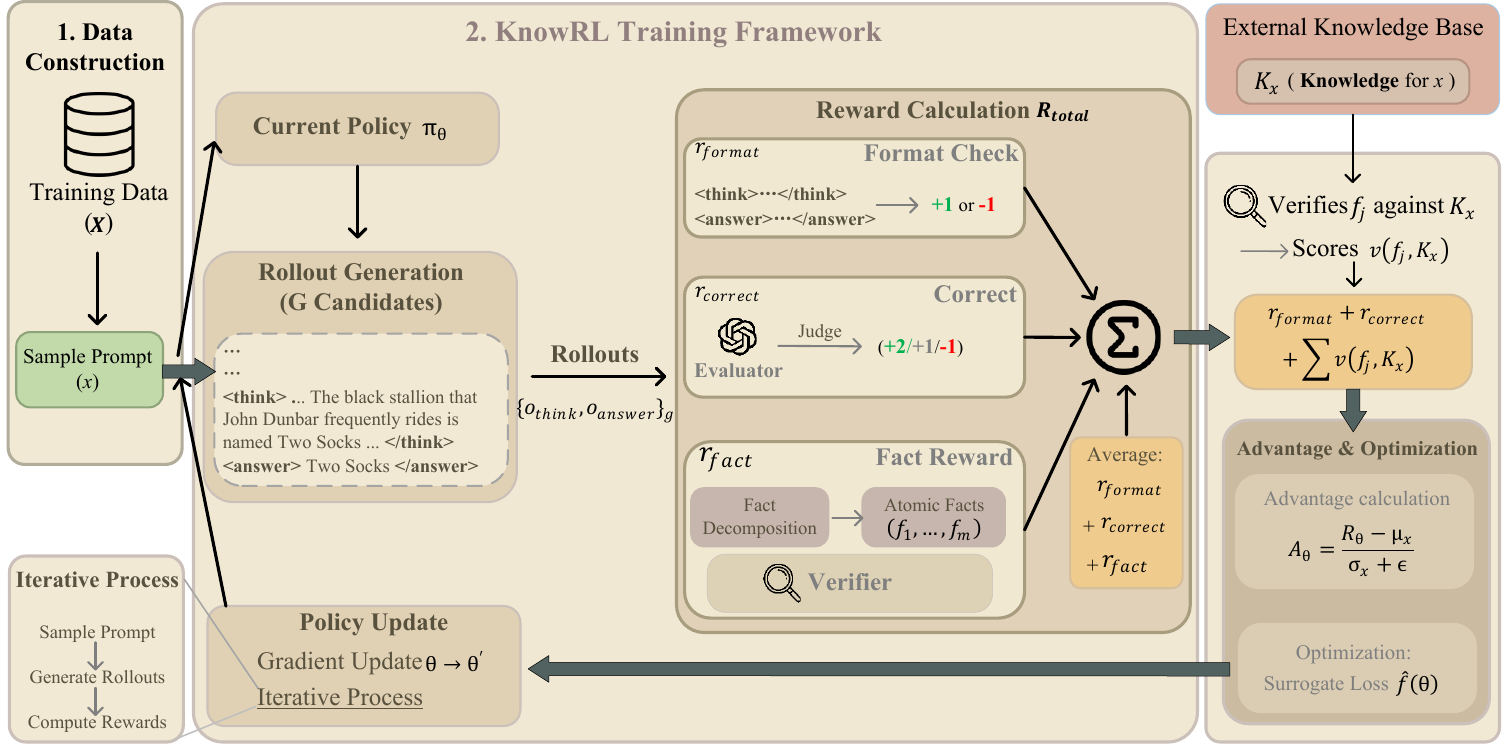}
    \caption{\textbf{Overview of the KnowRL framework.} We first construct training data matched with external knowledge. During RL training, we decompose the generated reasoning process into atomic facts and verify them against the knowledge base. Finally, we optimize the model using a composite reward that combines these factual scores with correctness and format checks.}
    \label{knowrl}
\end{figure*}

\section{Knowledgeable Reinforcement Learning}

To address the high hallucination rates in slow-thinking models, we propose a fundamental shift in supervision: focusing on the \textbf{\textit{reasoning process}} rather than only on final outcomes. 
To this end, we introduce \texttt{KnowRL}, a Knowledge enhanced Reinforcement Learning framework designed to guide models toward verifiable and boundary-aware reasoning by integrating factual supervision directly into the RL loop. 
As illustrated in Figure~\ref{knowrl}, we first construct training data matched with knowledge, and then conduct RL training using a composite reward that balances factual correctness with reasoning quality (details in Appendix~\ref{appendix:data-construction}). 
By enforcing factual consistency at the step level, KnowRL effectively corrects incorrect reasoning paths and encourages models to recognize their \textbf{\textit{knowledge boundaries}}, ultimately leading to a more faithful and self-aware thinking process.

KnowRL builds upon the Group Relative Policy Optimization (GRPO) algorithm by introducing \emph{factuality supervision} into the CoT reasoning process. This integration ensures that the model is guided not just by the final answer, but by the factual accuracy of its thinking steps. In the following, we detail how we design the composite Reward Function to evaluate these steps and how we implement the Factuality-Guided Policy Optimization.

\medskip
\noindent\textbf{Reward Function.}  
Given a rollout \(o=(o_{\mathrm{think}},o_{\mathrm{answer}})\), inspired by FactScore~\citep{min2023factscore}, we use GPT-4o-mini for the factual verification process. 
We decompose the reasoning trace into \(M\) atomic facts \(\Phi(o_{\mathrm{think}})=\{f_1,\dots,f_M\}\). 
Given an external knowledge base $K$, each atomic fact $f_j$ is checked against its most relevant knowledge set $K_x \subseteq K$\footnote{Retrieved using \texttt{sentence-transformers/gtr-t5-large}.} to obtain a verification score $v(f_j, K_x) \in \{0,1\}$.
The factuality reward is then defined as the proportion of supported facts:
\begin{equation}
\label{eq:factual-score}
r_{\mathrm{fact}}(o)=
\begin{cases}
\displaystyle \frac{1}{M}\sum_{j=1}^{M}v(f_j,K_x), & M>0,\\[6pt]
0, & M=0 .
\end{cases}
\end{equation}

Additional components include a \emph{\textbf{format reward}} \(r_{\mathrm{format}}(o)\) and a \emph{\textbf{correct reward}} \(r_{\mathrm{correct}}(o)\). 
The format reward verifies whether the output follows the required \texttt{<think>...</think><answer>...</answer>} structure: if the format is valid, \(r_{\mathrm{format}}(o)=+1\); otherwise, \(r_{\mathrm{format}}(o)=-1\). 
The correctness reward evaluates the final answer \(o_{\mathrm{answer}}\) using an evaluator model (GPT-4o-mini): if the answer is correct, \(r_{\mathrm{correct}}(o)=+2\); if the model explicitly refuses, \(r_{\mathrm{correct}}(o)=+1\); if the answer is incorrect, \(r_{\mathrm{correct}}(o)=-1\). 
The composite reward is then defined as
\begin{equation}
R_{\mathrm{total}}(o) = r_{\mathrm{format}}(o) + r_{\mathrm{correct}}(o) + r_{\mathrm{fact}}(o).
\end{equation}

\medskip

\noindent\textbf{Factuality–Guided Policy Optimisation.}\;
For every prompt $x$ the current policy
$\pi_{\theta_{\text{old}}}$ generates a \emph{group} of $G$ candidate
roll-outs
$\{o^{(g)}\}_{g=1}^{G}\!\sim\!\pi_{\theta_{\text{old}}}(\cdot|x)$, where
$G$ is the group size. 
Each trajectory is scored by the composite reward
$R_{\mathrm{total}}(\cdot)$, yielding a set of scalars
$\mathcal{R}_x=\{R_g\}_{g=1}^{G}$.

\noindent\textbf{Advantage construction.}
We summarise $\mathcal{R}_x$ with its sample mean
$\mu_{x}$ and standard deviation $\sigma_{x}$. 
The credit assigned to trajectory $g$ is the
\emph{group-relative advantage}
\begin{equation}
\label{eq:adv}
A_{g}\;=\;\frac{R_{g}-\mu_{x}}{\sigma_{x}+\varepsilon},
\end{equation}
where $\varepsilon\!(\!\ll\!1)$ avoids division by~zero.
$A_{g}$ is positive if $o^{(g)}$ attains above-average factual reward and negative otherwise, turning factual supervision into a signed learning signal.

\noindent\textbf{Likelihood ratio.}
Define the trajectory-level importance ratio
\[
\varrho_{g}
=\frac{\pi_{\theta}\!\bigl(o^{(g)}\!\mid x\bigr)}
       {\pi_{\theta_{\mathrm{old}}}\!\bigl(o^{(g)}\!\mid x\bigr)}.
\]
The pair $(\varrho_{g},A_{g})$ fully characterises how $o^{(g)}$ should
influence the update: \emph{increase} its probability if
$\varrho_{g}A_{g}>0$ and \emph{decrease} otherwise.

\noindent\textbf{Surrogate objective.}
Using these short symbols enables a compact, column-friendly surrogate:
\begin{equation}
\label{eq:surrogate}
\hat{\mathcal{J}}(\theta)=
\frac{1}{G}\sum_{g=1}^{G}
\min\!\bigl(
      \varrho_{g}A_{g},\,
      \operatorname{clip}(\varrho_{g},1-\epsilon,1+\epsilon)A_{g}
     \bigr),
\end{equation}
with a small clip threshold $\epsilon$ to bound the update.
Because $A_{g}$ encodes factual advantage,
maximising $\hat{\mathcal{J}}$ explicitly transfers probability mass from hallucination-heavy traces ($A_{g}<0$) to trajectories whose CoT is knowledge-supported ($A_{g}>0$).

\noindent\textbf{Regularised loss.}\;
To preserve exploration and limit divergence from the frozen reference policy $\pi_{\text{ref}}$, we add (i) an entropy bonus and (ii) a KL anchor.  
Let $o$ denote a complete rollout and $o_t$ its $t$-th token.  
The token-level Shannon entropy on prompt $x$ is therefore
$\mathcal{H}(\pi_{\theta}(\cdot|x))
      =-\sum_{t}\pi_{\theta}(o_t|x)\log\pi_{\theta}(o_t|x)$.
Its mini-batch expectation and the corresponding KL expectation are
$\mathcal{E}_{\mathcal{H}}\triangleq\mathbb{E}_{x}[\mathcal{H}(\pi_{\theta}(\cdot|x))]$ and
$\mathcal{E}_{\mathrm{KL}}\triangleq\mathbb{E}_{x}[D_{\mathrm{KL}}(\pi_{\theta}(\cdot|x)\|\pi_{\text{ref}}(\cdot|x)]$.
With scalar weights $\beta_{\mathcal{H}},\beta_{\mathrm{KL}}>0$, the final objective becomes
\begin{equation}
\label{eq:loss}
\mathcal{L}_{\text{KnowRL}}
  = -\hat{\mathcal{J}}(\theta)
    \;+\;\beta_{\mathcal{H}}\mathcal{E}_{\mathcal{H}}
    \;+\;\beta_{\mathrm{KL}}\mathcal{E}_{\mathrm{KL}}.
\end{equation}

During training, each gradient step proceeds as follows.  
First, a mini-batch of prompts $x$ is sampled and the old policy
$\pi_{\theta_{\mathrm{old}}}$ produces a group $\mathcal{O}_x$ of
$G$ roll-outs.
Next, every trajectory $o^{(g)}\!\in\!\mathcal{O}_x$ undergoes the
knowledge check that yields the composite reward
$R_g=R_{\mathrm{total}}(o^{(g)})$, whose factual component
$r_{\text{fact}}$ is computed against the external knowledge base.
The set $\{R_g\}$ is then converted into group-relative advantages
$A_g$ via Eq.\,\eqref{eq:adv}, so that a higher proportion of verified
facts directly translates into a larger positive $A_g$, whereas
hallucinated chains of thought receive negative credit.
The new policy $\pi_{\theta}$ is updated by maximising the factual
surrogate $\hat{\mathcal{J}}(\theta)$ in Eq.\,\eqref{eq:surrogate}
while minimising the regularised loss
$\mathcal{L}_{\text{KnowRL}}$ in Eq.\,\eqref{eq:loss};
this couples the likelihood ratio $\varrho_g$ with $A_g$ so that
gradients \emph{increase} the probability of
knowledge supported reasoning and \emph{decrease} that of
hallucination prone trajectories.  
Iterating this loop prompt sampling, factual verification,
advantage normalisation, and loss-driven update---yields a policy that
systematically suppresses hallucinations yet preserves answer accuracy.

\begin{table}[t]
    \centering
    \small
    \begin{tabular}{llc}
    \toprule
    \textbf{Sub-task} & \textbf{DeepSeek-7B} & \textbf{Skywork-7B} \\
    \midrule
    \multicolumn{3}{l}{\textit{Mathematics}} \\
    \quad COMP.en   & 8.61\% $\rightarrow$ 8.46\%  & 4.75\% $\rightarrow$ 7.12\% \\
    \quad COMP.zh   & 7.35\% $\rightarrow$ 7.60\%  & 5.39\% $\rightarrow$ 5.39\% \\
    \quad CEE.zh    & 10.65\% $\rightarrow$ 10.56\% & 8.79\% $\rightarrow$ 9.35\% \\
    \midrule
    \multicolumn{3}{l}{\textit{Physics}} \\
    \quad COMP.en   & 1.69\% $\rightarrow$ 0.00\%  & 0.85\% $\rightarrow$ 1.69\% \\
    \quad CEE.zh    & 7.83\% $\rightarrow$ 11.30\% & 3.48\% $\rightarrow$ 5.22\% \\
    \midrule
    \textbf{Average Acc.}  & \textbf{7.23\% $\rightarrow$ 7.58\%} & \textbf{4.65\% $\rightarrow$ 5.75\%} \\
    \bottomrule
    \end{tabular}
    \caption{\textbf{Performance on OlympiadBench.} Results are shown as Base model $\rightarrow$ KnowRL-trained.}
    \vspace{-3ex}
    \label{tab:olympiadbench_optimized_concise}
\end{table}

\begin{table*}[t]
    \centering
    \footnotesize 
    \setlength{\tabcolsep}{3.0pt} 
    \begin{tabular}{l|cc|cc@{\hskip 10pt}cc|cc@{\hskip 10pt}cc|cc}
    \toprule
    \multirow{3}{*}{\makecell[c]{\textbf{Methods}}} & \multicolumn{10}{c|}{\textbf{Hallucination}} & \multicolumn{2}{c}{\textbf{Reasoning}} \\
    \cmidrule(lr){2-11} \cmidrule(lr){12-13}
    & \multicolumn{2}{c|}{\textbf{TruthfulQA}} & \multicolumn{4}{c|}{\textbf{SimpleQA}} & \multicolumn{4}{c|}{\textbf{ChineseSimpleQA}} & \multirow{2}{*}{\makecell[c]{\small\textbf{GPQA}\\\small\textbf{Diamond}}} & \multirow{2}{*}{\small\textbf{AIME}}\\
    \cmidrule(lr){2-3} \cmidrule(lr){4-7} \cmidrule(lr){8-11}
    & \small\textbf{Rouge} & \small\textbf{Bleu} & \small\textbf{PAQ} & \small\textbf{Incorrect} & \small\textbf{Refusal} & \small\textbf{F1} & \small\textbf{PAQ} & \small\textbf{Incorrect} & \small\textbf{Refusal} & \small\textbf{F1} & & \\
    \midrule
    \rowcolor{lightgray} \multicolumn{13}{c}{\textit{\textbf{Skywork-OR1-7B-Preview}}} \\
    \midrule
    \text{Zero-shot}    & 56.67 & \textbf{55.33} & 2.97 & 76.33 & 21.33 & 2.61 & 11.84 & 67.00 & 24.00 & 10.23 & 37.37 & 26.67  \\
    \text{Self-Refine}   & \underline{58.00} & 54.00 & 3.90 & \perf{74.00}{-2.33} & 23.00 & 3.90 & 9.82 & \perf{67.33}{+0.33} & 25.33 & 8.40 & 46.67 & 36.67 \\
    \text{FactTune-FS}       & \textbf{58.33} & 50.33 & 0.76 & \perf{43.33}{-33.0} & \textbf{56.33} & 0.46 & 8.52 & \perf{68.00}{+1.00} & \underline{25.67} & 7.26 & 40.91 & 26.67  \\
    \text{DPO}                 & 52.67 & 49.33 & \underline{4.81} & \perf{85.66}{+9.33} & 10.00 & \underline{4.56} & \underline{12.64} & \perf{78.33}{+11.3} & 10.33 & \underline{11.95} & 34.85 & 30.00 \\
    \text{SFT}                 & 57.67 & 51.67 & \textbf{11.45} & \perf{77.33}{+1.00} & 12.67& \textbf{10.68} & \textbf{19.70}&\perf{70.67}{+3.67}& 12.00& \textbf{18.44}& 34.85 & 23.33  \\
     \text{TruthRL}              & 57.33 & 50.60 & 3.78 & \perf{56.00}{-20.3} & \underline{41.67} & 2.95 & 10.58 & \perf{62.00}{-5.00} & 30.67 & 8.66 & 39.39 & 26.67 \\
    \text{KnowRL}                & 57.67 & \underline{54.33} & 3.21 & \perf{60.33}{-16.0} & 37.67 & 2.46 & 12.29 & \perf{52.33}{-14.7} & \textbf{40.33} & 9.19 & 42.42 $\uparrow$ & 36.67 $\uparrow$ \\[3pt]
    \midrule
    \rowcolor{lightgray} \multicolumn{13}{c}{\textit{\textbf{DeepSeek-R1-Distill-Qwen-7B}}} \\
    \midrule
    \text{Zero-shot}    & 53.33 & 51.00 & 2.09 & 78.00 & 20.33 & 1.86 & 8.07 & 68.33 & 25.67 & 6.88 & 40.91 & 30.00  \\
    \text{Self-Refine}         & 55.00 & 50.33 & 2.52 & \perf{77.33}{-0.67} & 20.67 & 2.23 & 8.11 & \perf{68.00}{-0.33} & \underline{26.00} & 6.90 & 45.45 & 33.33  \\
    \text{FactTune-FS}         & 54.00 & 50.00 & 2.72 & \perf{59.67}{-18.3} & \underline{38.67} & 2.07 & 10.24 & \perf{76.00}{+7.67} & 15.33 & 9.39 & 38.89 & 30.00  \\
    \text{DPO}                 & 54.00 & 51.00 & \underline{4.35} & \perf{88.00}{+10.0} & 8.00 & \underline{4.16} & \underline{13.14} & \perf{79.33}{+11.0} & 8.67 & \underline{12.54} & 37.37 & 30.00 \\
    \text{SFT}                 & \textbf{57.67} & \underline{52.00} & \textbf{8.42} & \perf{83.33}{+5.33}& 9.00 & \textbf{8.03} & \textbf{19.58} & \perf{76.67}{+8.34}& 4.67& \textbf{19.11} & 36.36 & 26.67  \\
    \text{TruthRL}              & \textbf{57.67} & \textbf{54.67} & 2.14 & \perf{61.00}{-17.0} & 37.67 & 1.64 & 5.76 & \perf{60.00}{-8.33} & \textbf{36.33} & 4.48 & 39.39 & 26.67 \\
    \text{KnowRL}                & \underline{57.33} & 51.60 & 2.81 & \perf{57.67}{-20.3} & \textbf{40.67} & 2.09 & 10.26 & \perf{58.33}{-10.0} & 35.00 & 8.08 & 36.87 $\downarrow$ &  33.33 $\uparrow$ \\
    \bottomrule
    \end{tabular}
    \caption{\textbf{Main Experimental Results.} Performance of KnowRL compared against baselines on the Skywork-OR1-7B-Preview and DeepSeek-R1-Distill-Qwen-7B models. Zero-shot refers to the original model performance. The best results are marked in \textbf{bold}.}
    \vspace{-3ex}
    \label{main-exp}
\end{table*}

\begin{figure*}[t]
    \centering
    \begin{subfigure}[b]{0.32\textwidth}
        \centering
        \includegraphics[width=\linewidth]{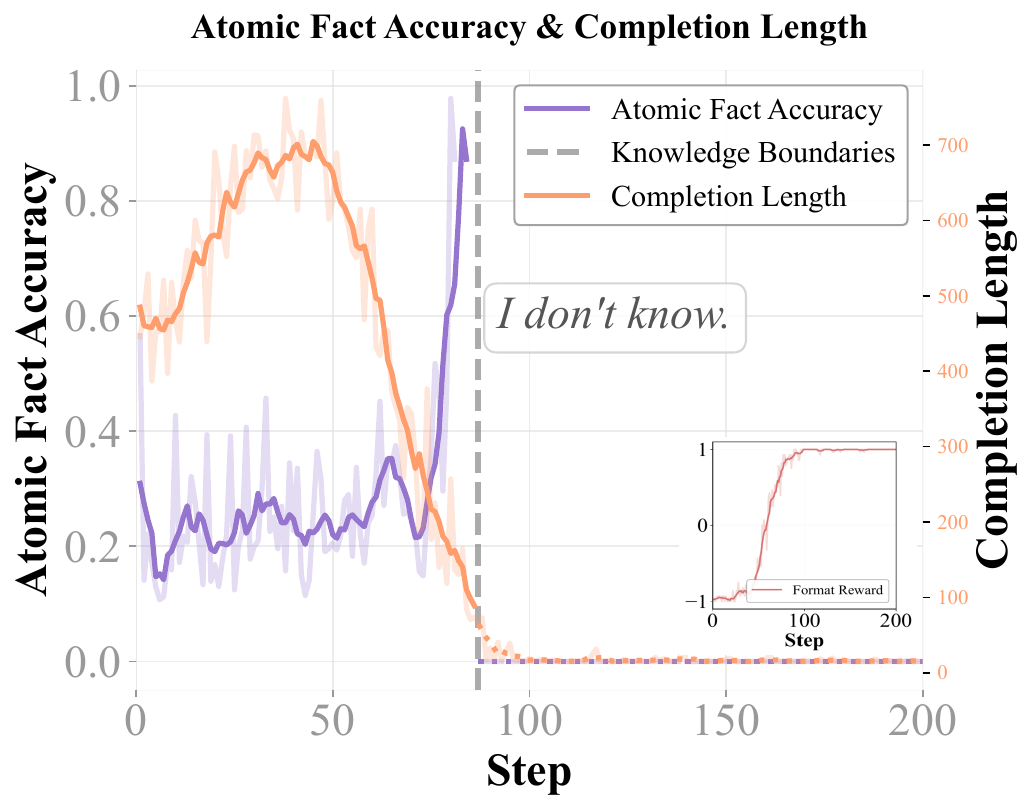}
        \caption{The Training Curves (Skywork-OR1-7B-Preview)}
        \label{fig:sub1}
    \end{subfigure}
    \hfill
    \begin{subfigure}[b]{0.32\textwidth}
        \centering
        \includegraphics[width=\linewidth]{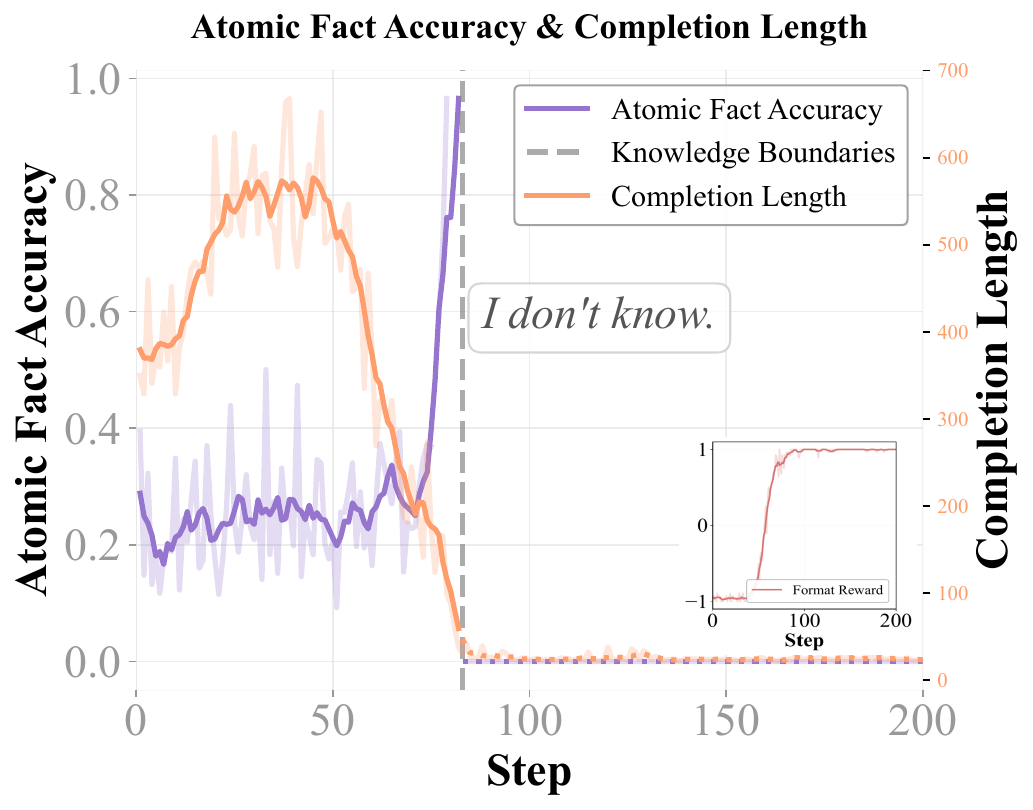}
        \caption{The Training Curves (DeepSeek-R1-Distill-Qwen-7B)}
        \label{fig:sub2}
    \end{subfigure}
    \hfill
    \begin{subfigure}[b]{0.32\textwidth}
        \centering
        \includegraphics[width=\linewidth]{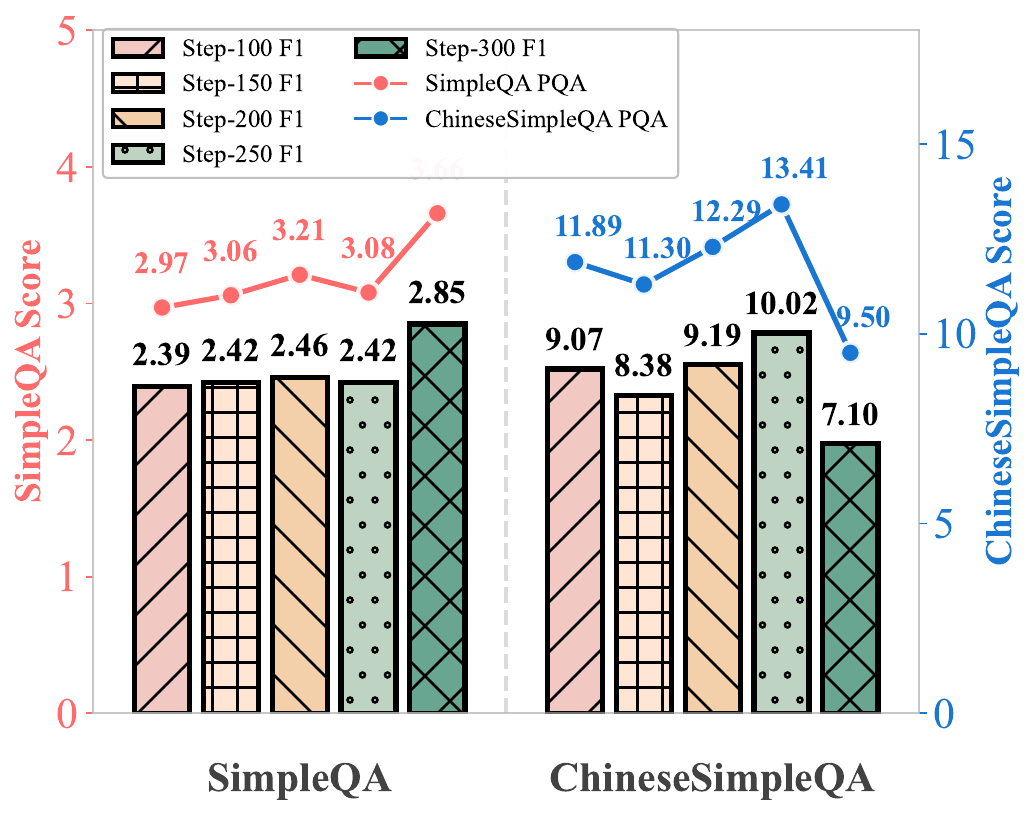}
        \caption{F1 and PAQ Across Training Steps on SimpleQA and ChineseSimpleQA}
        \label{fig:sub3}
    \end{subfigure}
    \hfill
    \begin{subfigure}[b]{0.32\textwidth}
        \centering
        \includegraphics[width=\linewidth]{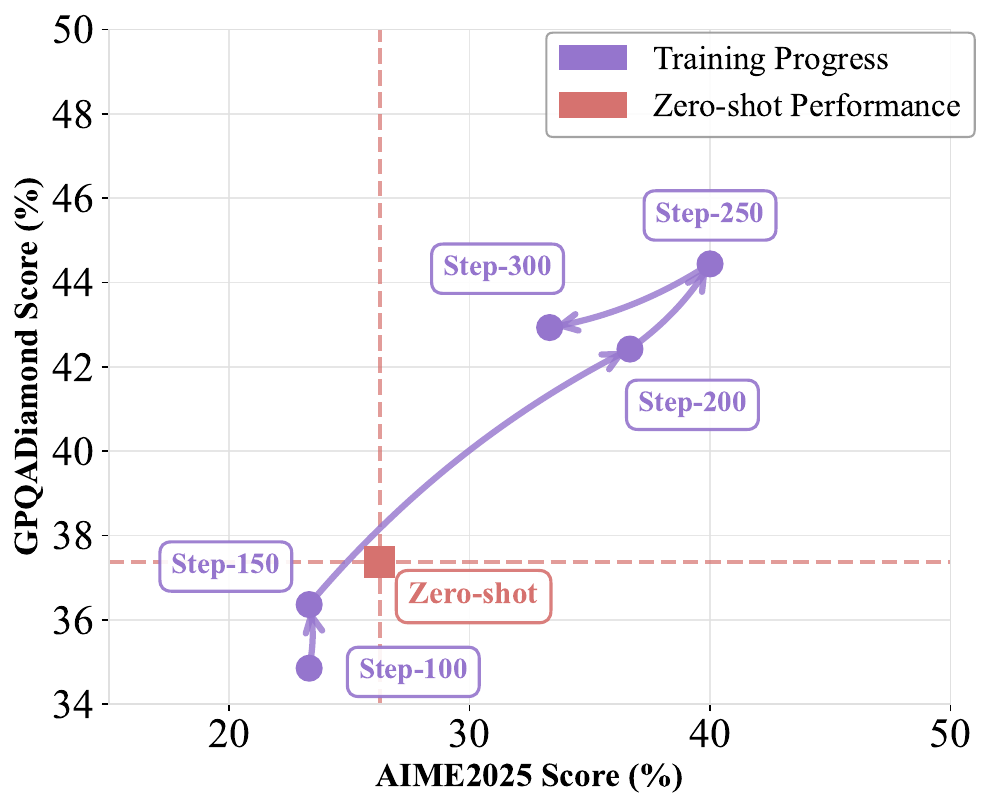}
        \caption{Training Progress on GPQA and AIME~2025}
        \label{fig:sub4}
    \end{subfigure}
    \hfill
    \begin{subfigure}[b]{0.32\textwidth}
        \centering
        \includegraphics[width=\linewidth]{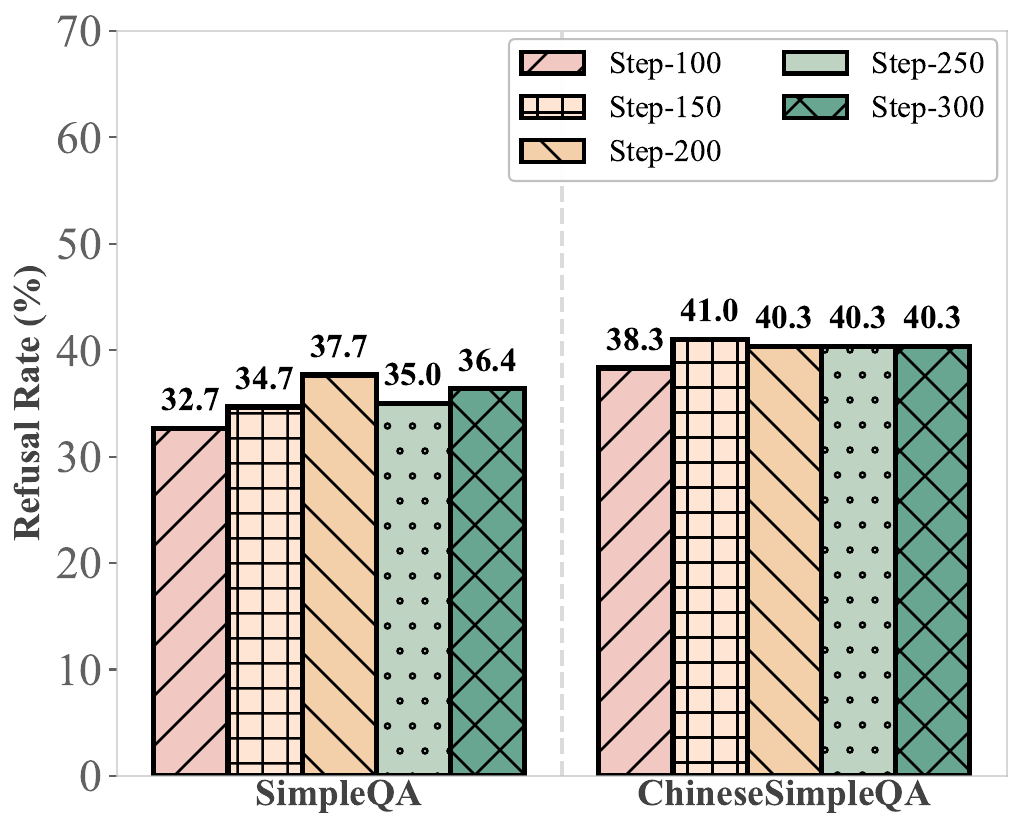}
        \caption{Refusal Rates Across Training Steps on SimpleQA and ChineseSimpleQA}
        \label{fig:sub5}
    \end{subfigure}
    \hfill
    \begin{subfigure}[b]{0.32\textwidth}
        \centering
        \includegraphics[width=\linewidth]{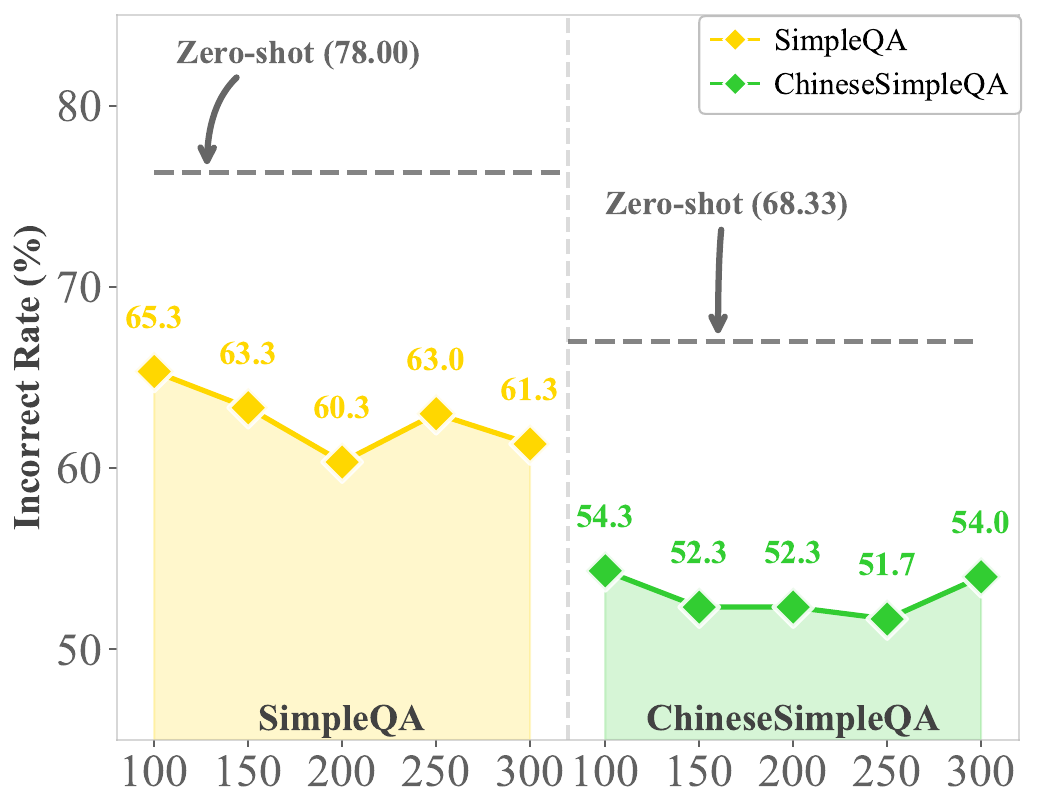}
        \caption{Incorrect Rates Across Training Steps vs. Zero-shot Performance}
        \label{fig:sub6}
    \end{subfigure}
    \caption{
    KnowRL training dynamics and reasoning behavior analysis. 
    (a)--(b) present the training curves of two model architectures, showing improved factual alignment and stable reasoning length. 
    (c)--(f) illustrate performance trends of Skywork-OR1-7B-Preview across different training steps, including F1 and PAQ (c), reasoning accuracy (d), refusal rate (e), and incorrect rate (f).
    }
    \vspace{-3ex}
    \label{fig:knowrl-exp}
\end{figure*}

\section{Experiments}
\subsection{Experimental Settings.}
\paragraph{Datasets and Metrics.}
We use TruthfulQA~\citep{lin2021truthfulqa}, SimpleQA, and ChineseSimpleQA to evaluate hallucination, and GPQA (general domain) and AIME 2025 (mathematical reasoning) to evaluate reasoning ability.
For the hallucination evaluation datasets, we randomly sample 300 examples from each of TruthfulQA, SimpleQA, and ChineseSimpleQA as test sets.
Because slow-thinking models generate thousands of tokens per query, evaluating full datasets is computationally prohibitive.
A 300-example subset remains highly challenging and provides a statistically valid evaluation, which is a standard practice for long-reasoning tasks.
TruthfulQA is assessed with the ROUGE and BLEU metrics.
For SimpleQA and ChineseSimpleQA, we evaluate performance using four metrics: (1) Incorrect Rate, (2) Refusal Rate, (3) Precision on Answered Questions (PAQ), and (4) $F_1$ score.
PAQ measures the proportion of correctly answered questions among those that the model chooses to answer.
For AIME 2025 and GPQA, we assess reasoning performance based on accuracy metric.
All models are evaluated with the temperature set to 0.

\paragraph{Models and Baselines.}
We select the Skywork-OR1-7B-Preview \citep{skywork-or1-2025} and the DeepSeek-R1-Distill-Qwen-7B for experiments. 
These models represent the two most popular slow thinking training methods: RL, represented by the Skywork-OR1-7B-Preview model, and distillation, represented by the DeepSeek-R1-Distill-Qwen-7B model. 
We select Self-Refine \citep{madaan2023self} as the baseline for prompt engineering, and SFT, DPO, FactTune-FS \citep{tian2023fine} and TruthRL \citep{wei2025truthrl} as the baselines for post-training methods. 
For DPO, we use the distilled DeepSeek-R1 data as the chosen data; for FactTune-FS, the chosen data is composed of the original model's outputs, which are filtered for high factuality using FactScore. 
We use Low-Rank Adaptation \citep[LoRA;][]{hu2022lora} to train all models. Further training details are provided in the Appendix~\ref{appendix:train-set}.


\subsection{Main Results}
\paragraph{KnowRL significantly mitigates hallucination while maintaining reasoning capabilities.}
As shown in Table~\ref{main-exp}, KnowRL consistently improves factual reliability across all datasets. 
For the DeepSeek-R1-Distill-Qwen-7B model, the Incorrect Rate on SimpleQA drops by over 20\% (from 78.00\% to 57.67\%), with similar improvements observed on ChineseSimpleQA. 
Notably, since our training knowledge source is primarily English-based, the consistent gains on ChineseSimpleQA highlight strong cross-lingual transferability, suggesting that KnowRL helps the model internalize a language-agnostic verification behavior rather than merely memorizing language-specific factual patterns.
Importantly, unlike baseline methods that often suffer from catastrophic forgetting, KnowRL preserves strong performance on complex reasoning benchmarks. 
For instance, it improves the GPQA score from 37.37\% to 42.42\% and maintains stability on AIME 2025, demonstrating that factual supervision does not conflict with deep reasoning abilities.
This favorable trade-off can be attributed to KnowRL's process-level design: by rewarding the factual grounding of individual reasoning steps rather than solely adjusting outcome distributions, the model retains its capacity for extended logical deduction while learning to ground each step in verifiable knowledge.
Furthermore, the dramatic drop in completion length (Figure~\ref{fig:knowrl-exp}(a) and (b)) reflects acquired boundary-awareness rather than a collapse of the slow-thinking process.
The model efficiently truncates wasteful hallucinations for unknown facts while preserving extended reasoning chains for complex logical tasks.

\paragraph{Training dynamics demonstrate that the model progressively learns to recognize knowledge boundaries.}
Figure~\ref{fig:knowrl-exp}(a) and (b) illustrates how KnowRL shapes the model's behavior. 
During training, the accuracy of atomic facts within the chain-of-thought steadily increases. 
As the model becomes more precise in its reasoning steps, it learns to distinguish known facts from unknown information. 
Consequently, instead of fabricating answers when uncertain, the model learns to abstain, reflected by a rise in the Refusal Rate and a sharp drop in the Incorrect Rate. 
This confirms that KnowRL effectively teaches the model to reason faithfully rather than merely optimizing for lucky guesses.

\begin{table*}[h]
    \centering
    \footnotesize 
    \setlength{\tabcolsep}{3.0pt} 
    \begin{tabular}{l|cc@{\hskip 14pt}cc|cc@{\hskip 14pt}cc|cc}
    \toprule
    \multirow{3}{*}{\makecell[c]{\textbf{Method}}} & \multicolumn{8}{c|}{\textbf{Hallucination}} & \multicolumn{2}{c}{\textbf{Reasoning}} \\
    \cmidrule(lr){2-9} \cmidrule(lr){10-11}
    & \multicolumn{4}{c|}{\textbf{SimpleQA}} & \multicolumn{4}{c|}{\textbf{ChineseSimpleQA}} & \multirow{2}{*}{\makecell[c]{\small\textbf{GPQA}\\\small\textbf{Diamond}}} & \multirow{2}{*}{\small\textbf{AIME}}\\
    \cmidrule(lr){2-5} \cmidrule(lr){6-9}
    & \small\textbf{PAQ} & \small\textbf{Incorrect} & \small\textbf{Refusal} & \small\textbf{F1} & \small\textbf{PAQ} & \small\textbf{Incorrect} & \small\textbf{Refusal} & \small\textbf{F1} & & \\
    
    \midrule
    \rowcolor{lightgray} \multicolumn{11}{c}{\textit{\textbf{Ablation on Reward Combination (Base Model: DeepSeek-R1-Distill-Qwen-7B)}}} \\
    \midrule
    $R = r_{\text{format}}$ & 2.63 & 74.00 & 24.00 & \underline{2.27} & 7.08 & 74.33 & 20.00 & 6.29 & \underline{39.39} & 30.00  \\
    $R = r_{\text{format}} + r_{\text{fact}}$ & 2.42 & \perf{80.67}{+6.67} & 17.33 & 2.19 & 8.87 & \perf{75.33}{+1.00} & 17.33 & \underline{8.03} & \textbf{47.47} & \textbf{40.00} \\
    $R = r_{\text{format}} + r_{\text{correct}}$ & \textbf{3.19} & \perf{60.67}{-13.33} & \underline{37.33} & \textbf{2.46} & \underline{8.89} & \perf{41.00}{-33.33} & \textbf{55.00} & 5.52 & 38.89 & \textbf{40.00} \\
    $R = R_{total}$ (KnowRL) & \underline{2.81} & \perf{57.67}{-16.33} & \textbf{40.67} & 2.09 & \textbf{10.26} & \perf{58.33}{-16.00} & \underline{35.00} & \textbf{8.08} & 36.87 & \underline{33.33} \\
    \addlinespace[0.5pt] 
    KnowRL ($R_{refusal}=-1$) & 1.26 & \perf{78.67}{+4.67} & 20.33 & 1.11 & 7.66 & \perf{84.33}{+10.00} & 8.67 & 7.32 & 34.85 & 30.00 \\
    
    \addlinespace[2pt]
    \midrule
    \rowcolor{lightgray} \multicolumn{11}{c}{\textit{\textbf{Robustness Analysis (Using GRPO reward, $R = R_{total}$)}}}\\
    \midrule
    
    Zero-shot & 2.09 & 78.00 & 20.33 & 1.86 & 8.07 & 68.33 & 25.67 & 6.88 & 40.91 & 30.00 \\
    \addlinespace[0.5pt]
    KnowRL (DAPO) & 1.11 & \perf{59.33}{-18.67} & 40.00 & 0.83 & 9.14 & \perf{59.67}{-8.66} & 34.33 & 7.24 & 41.41 & 43.33 \\
    KnowRL (BNPO) & 1.61 & \perf{61.00}{-17.00} & 38.00 & 1.23 & 11.48 & \perf{61.67}{-6.66} & 30.33 & 9.43 & 41.41 & 30.00 \\
    KnowRL (Dr.GRPO) & 3.16 & \perf{61.33}{-16.67} & 36.67 & 2.45 & 11.33 & \perf{60.00}{-8.33} & 32.33 & 9.15 & 38.38 & 43.33 \\

    \bottomrule
    \end{tabular}
    \caption{\textbf{Ablation and Robustness Analysis.} Impact of reward components and different RL algorithms for the DeepSeek-R1-Distill-Qwen-7B. In reward ablation section, best results are in \textbf{bold} and second-best are \underline{underlined}.}
    \vspace{-3ex}
    \label{tab:ablation}
\end{table*}

\subsection{Extended Evaluation on Reasoning Benchmarks}
We further evaluate KnowRL on OlympiadBench~\citep{he2024olympiadbench} to examine its impact on complex reasoning in mathematics and physics. 
As shown in Table~\ref{tab:olympiadbench_optimized_concise}, both models maintain or improve their performance after KnowRL training. 
Specifically, the DeepSeek model increases its average accuracy from 7.23\% to 7.58\%, while the Skywork model improves from 4.65\% to 5.75\%. 
Both models achieved significant gains in the Chinese physics task (\textit{physics\_zh\_CEE}), with DeepSeek-7B rising from 7.83\% to 11.30\%. 
Although there are minor fluctuations in individual sub-tasks, the overall trend confirms that factual supervision does not hinder structured reasoning.

Combined with the consistent results on GPQA and AIME~2025, these findings provide strong evidence that \textbf{KnowRL effectively preserves and enhances complex reasoning capabilities across diverse other domains, enabling models to reason more reliably under factual constraints}.

\subsection{Training Step Analysis}
To investigate how the number of KnowRL training steps affects the trade-off between factuality and reasoning, we evaluate models trained for 100, 150, 200, 250, and 300 reinforcement learning steps. 
The results are illustrated in Figure~\ref{fig:knowrl-exp}(c)--(f).

As shown in Figure~\ref{fig:knowrl-exp}(c), both F1 and PAQ scores improve rapidly in the early stage of training and reach a stable high level after moderate training, suggesting that factual precision and reliability are primarily acquired in the middle phase of optimization. 
Reasoning performance on GPQA and AIME~2025 (Figure~\ref{fig:knowrl-exp}(d)) remains stable or slightly improves within this range, indicating that factual supervision can refine reasoning without inducing degradation. 
Refusal rates (Figure~\ref{fig:knowrl-exp}(e)) increase and then plateau, reflecting that models progressively learn appropriate abstention behavior, while the incorrect rate (Figure~\ref{fig:knowrl-exp}(f)) decreases sharply in early training and stabilizes thereafter. 
Beyond this stage, additional optimization provides limited factual improvement and introduces minor task-dependent variations rather than consistent gains.

These observations suggest that KnowRL’s reinforcement learning dynamics follow a stable convergence pattern rather than the reversal effects sometimes observed in overfitted reasoning models~\citep{berglund2023reversal}. 
Appropriate training duration is therefore crucial: insufficient updates hinder factual learning, whereas excessive optimization may overfit factual supervision signals and weaken generalization.

\textbf{Overall, KnowRL achieves its best balance between factuality and reasoning stability when trained for an appropriate number of reinforcement learning steps, where the model fully internalizes factual supervision without over-optimization or reasoning drift.}

\subsection{Ablation Study}


To understand how different reward signals contribute to the model's performance, we evaluate the DeepSeek-R1-Distill-Qwen-7B model under different reward combinations.

\paragraph{Factual reward ($r_{fact}$) enhances fact-grounded reasoning.}
As shown in Table~\ref{tab:ablation}, using the factual reward alone achieves the best performance on reasoning benchmarks such as GPQA and AIME~2025. 
This shows that $r_{\text{fact}}$ encourages models to perform reasoning grounded in verifiable knowledge, helping them maintain accuracy while reducing random errors. 
This finding underscores the importance of factual verification in the reinforcement learning process, consistent with prior work showing that incorporating verifiable rewards can substantially improve the reliability and stability of RL training~\citep[RLVR;][]{yue2025does}.

\paragraph{Positive refusal incentives promote boundary awareness.} Table~\ref{tab:ablation} shows that giving a positive reward for refusals leads to higher refusal rates and fewer hallucination errors. 
This explicitly encourages the model to admit when it lacks knowledge.
When we removed this positive incentive (setting $R_{\text{refusal}}=-1$), the incorrect rate increased significantly.
This confirms that rewarding appropriate refusals is essential for learning knowledge boundaries.

\section{Analysis}

\subsection{Robustness Analysis of Different RL Algorithms}

To verify that the effectiveness of KnowRL is not limited to a specific optimization method, we extended our framework to three additional RL algorithms: BNPO \cite{xiao2025bnpo}, DAPO \citep{yu2025dapo}, and Dr.GRPO \citep{liu2025understanding}. 
We compare their performance using the DeepSeek-R1-Distill-Qwen-7B model.

\paragraph{KnowRL is robust across different algorithms.} 
Table~\ref{tab:ablation} demonstrates that every algorithm achieved the dual goal of our framework: they all significantly reduced the Incorrect Rate compared to the Zero-shot baseline, while successfully maintaining or even improving reasoning performance. 
This confirms that the KnowRL framework effectively reduces hallucinations without compromising reasoning capabilities, regardless of the specific RL algorithm used. 
Detailed analysis of response length and atomic fact accuracy during training is provided in Appendix~\ref{appendix:RL_algorithms}.

\subsection{Analysis of Factors Affecting Knowledge Boundaries}

To examine whether the positive refusal reward in $r_{\text{correct}}$ is the only factor driving boundary learning, we conducted an experiment on the DeepSeek-R1-Distill-Qwen-7B model using SimpleQA and ChineseSimpleQA. 
We modified the KnowRL setting by changing the refusal reward from positive to negative $r_{\text{refusal}=-1}$, while keeping other rewards unchanged. 
We evaluated the model at training steps 0, 100, 150, 200, and 250.

\begin{figure}[t]
    \centering
    \resizebox{.41\textwidth}{!}{
    \includegraphics{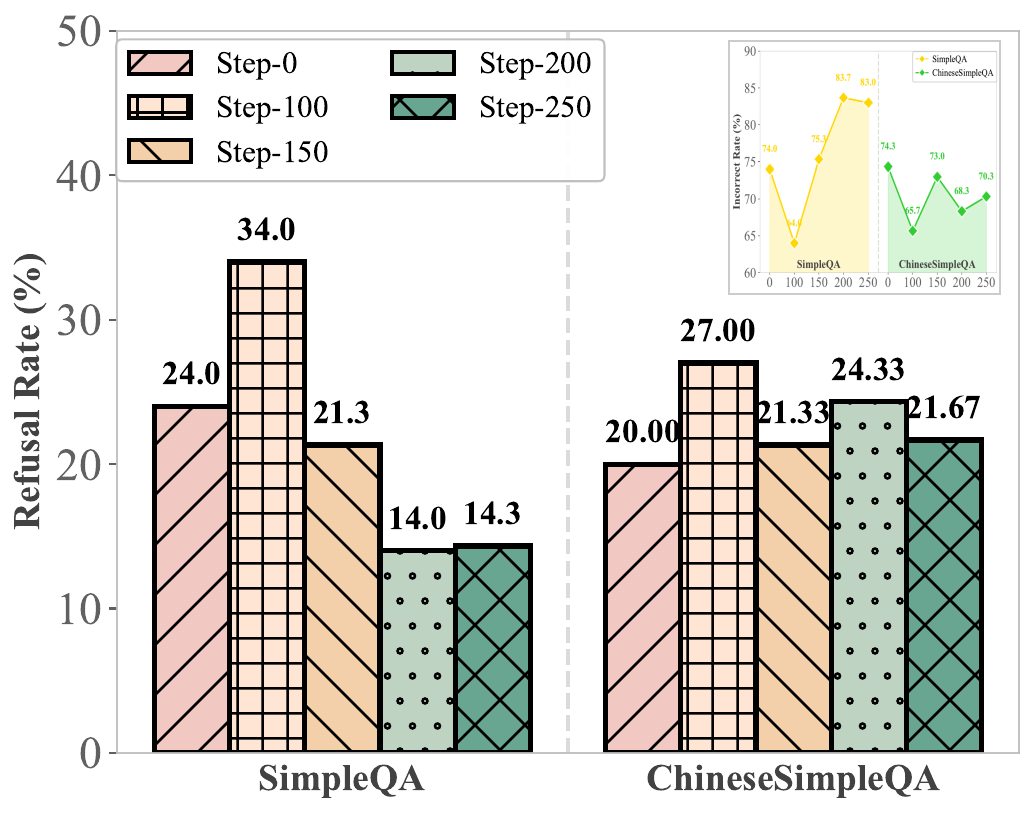}}
    \caption{
    Trends of refusal and incorrect rates on SimpleQA and ChineseSimpleQA 
    when training with the combination of format reward and negative refusal reward across different training steps.}
    \vspace{-3ex}
    \label{fig:boundary-analysis}
\end{figure}

As shown in Figure~\ref{fig:boundary-analysis}, the refusal rate initially rises even without a positive refusal reward. 

This confirms that the positive refusal reward is not the only factor affecting knowledge boundaries; the penalty for incorrect answers also encourages the model to be cautious in the early stages. 
However, this behavior is unstable. 
Later stages in Figure~\ref{fig:boundary-analysis} show a significant drop in refusal rates accompanied by a spike in incorrect answers. 
This occurs because the model learns to guess to maximize scores rather than accepting refusal penalties—known as reward hacking. Thus, while correctness rewards trigger initial caution, positive refusal rewards are essential to maintain it and prevent guessing.
\textbf{Overall, the results suggest that positive refusal rewards are not the only source of boundary learning; 
different reward combinations can jointly promote boundary-aware reasoning.}

We further investigate the scalability and evaluator sensitivity of our framework. 
Experiments on the larger DeepSeek-R1-Distill-Qwen-14B model show that KnowRL consistently reduces hallucinations while improving reasoning capabilities, confirming its scalability across model sizes(details in Appendix~\ref{appendix:scalability}). 
Moreover, we conduct an Evaluator Sensitivity Analysis by replacing the GPT-4o-mini used during training with different model. 
The results indicate that KnowRL maintains robust performance regardless of the specific evaluator used. Comprehensive results for these analyses are provided in Appendix~\ref{appendix:sensitivity}.

\section{Robustness and Multi-run Evaluation}
\label{app:multi_run}

To address the potential sensitivity of single-run evaluations at zero temperature, especially on small-scale benchmarks like AIME, we conduct a robust multi-run evaluation.
We re-evaluate the base models and KnowRL using a non-zero temperature of 0.6, generating 5 responses per prompt to report the average performance (Avg@5).
This setup ensures that the observed improvements in factuality and reasoning are consistent and not artifacts of decoding variance.

\begin{table*}[t]
    \centering
    \footnotesize 
    \setlength{\tabcolsep}{3.0pt} 
    \begin{tabular}{l|cc|cc@{\hskip 10pt}cc|cc@{\hskip 10pt}cc|cc}
    \toprule
    \multirow{3}{*}{\makecell[c]{\textbf{Methods}}} & \multicolumn{10}{c|}{\textbf{Hallucination}} & \multicolumn{2}{c}{\textbf{Reasoning}} \\
    \cmidrule(lr){2-11} \cmidrule(lr){12-13}
    & \multicolumn{2}{c|}{\textbf{TruthfulQA}} & \multicolumn{4}{c|}{\textbf{SimpleQA}} & \multicolumn{4}{c|}{\textbf{ChineseSimpleQA}} & \multirow{2}{*}{\makecell[c]{\small\textbf{GPQA}\\\small\textbf{Diamond}}} & \multirow{2}{*}{\small\textbf{AIME}}\\
    \cmidrule(lr){2-3} \cmidrule(lr){4-7} \cmidrule(lr){8-11}
    & \small\textbf{Rouge} & \small\textbf{Bleu} & \small\textbf{PAQ} & \small\textbf{Incorrect} & \small\textbf{Refusal} & \small\textbf{F1} & \small\textbf{PAQ} & \small\textbf{Incorrect} & \small\textbf{Refusal} & \small\textbf{F1} & & \\
    \midrule
    \rowcolor{lightgray} \multicolumn{13}{c}{\textit{\textbf{Skywork-OR1-7B-Preview}}} \\
    \midrule
    \text{Zero-shot}    & 56.20 & 51.60 & 7.48 & 59.47 & 35.73 & 9.16 & 10.95 & 57.87 & 35.00 & 13.30 & 44.95 & 34.67  \\
    \text{DPO}          & 56.13 & 51.33 & 6.65 & \perf{81.27}{+21.80} & 12.93 & 10.93 & 11.83 & \perf{77.47}{+19.60} & 12.13 & 18.81 & 44.34 & 34.67 \\
    \text{FactTune-FS}  & 57.60 & 53.00 & 3.51 & \perf{44.33}{-15.14} & 54.07 & 2.20 & 10.84 & \perf{40.60}{-17.27} & 54.47 & 9.40 & 45.45 & 34.67  \\
    \text{SFT}          & 57.27 & 50.93 & 9.52 & \perf{61.07}{+1.60} & 32.40 & 10.71 & 8.21 & \perf{78.93}{+21.06} & 14.00 & 7.59 & 39.39 & 38.67  \\
    \text{TruthRL}      & 57.67 & 53.47 & 8.15 & \perf{49.07}{-10.40} & 46.60 & 8.30 & 12.33 & \perf{32.60}{-25.27} & 62.80 & 8.78 & 46.36 & 33.33 \\
    \text{KnowRL}       & 58.27 & 52.80 & 7.32 & \perf{40.20}{-19.27} & \textbf{56.60} & 6.17 & 10.30 & \perf{28.40}{-29.47} & \textbf{68.33} & 6.32 & \textbf{48.89} $\uparrow$ & \textbf{38.00} $\uparrow$ \\[3pt]
    \midrule
    \rowcolor{lightgray} \multicolumn{13}{c}{\textit{\textbf{DeepSeek-R1-Distill-Qwen-7B}}} \\
    \midrule
    \text{Zero-shot}    & 57.40 & 51.53 & 4.89 & 62.47 & 34.33 & 6.20 & 10.23 & 60.27 & 32.93 & 11.83 & 45.45 & 29.33  \\
    \text{DPO}          & 55.60 & 51.40 & 5.65 & \perf{76.33}{+13.86} & 19.07 & 8.74 & 9.54 & \perf{67.20}{+6.93} & 25.73 & 13.20 & 40.40 & 28.67 \\
    \text{FactTune-FS}  & 56.93 & 52.13 & 2.73 & \perf{61.73}{-0.74} & 36.53 & 2.12 & 8.90 & \perf{56.60}{-3.67} & 37.87 & 10.48 & 43.74 & 30.67  \\
    \text{SFT}          & 57.13 & 51.27 & 9.38 & \perf{82.47}{+20.00} & 9.00 & 15.72 & 9.86 & \perf{82.07}{+21.80} & 8.93 & 9.41 & 38.28 & 35.33  \\
    \text{TruthRL}      & 56.27 & 51.94 & 5.93 & \perf{50.73}{-11.74} & 46.07 & 6.19 & 9.37 & \perf{30.47}{-29.80} & 66.40 & 4.69 & 44.34 & 31.33 \\
    \text{KnowRL}       & 57.20 & 52.40 & 5.50 & \perf{48.27}{-14.20} & \textbf{48.93} & 5.43 & 8.38 & \perf{28.07}{-32.20} & \textbf{69.40} & 4.94 & \textbf{46.97} $\uparrow$ & \textbf{34.00} $\uparrow$ \\
    \bottomrule
    \end{tabular}
    \caption{\textbf{Multi-run Evaluation Results.} Robust multi-run performance (Avg@5, Temperature=0.6) of KnowRL compared against baselines on the Skywork-OR1-7B-Preview and DeepSeek-R1-Distill-Qwen-7B models. Zero-shot refers to the original model performance. The key improvements are marked in \textbf{bold}.}
    \vspace{-3ex}
    \label{tab:multi_run_results}
\end{table*}

As shown in Table \ref{tab:multi_run_results}, the multi-run averages consistently support our primary findings.
Even under stochastic decoding, KnowRL significantly improves refusal rates on hallucination-prone queries (SimpleQA and ChineseSimpleQA) while simultaneously enhancing or maintaining performance on complex reasoning tasks (GPQA and AIME).
For instance, KnowRL improves AIME accuracy from 29.33\% to 34.00\% for DeepSeek-7B, confirming that our framework effectively promotes reliable reasoning rather than simply collapsing to conservative refusals.

\section{Related Work}
\paragraph{Hallucination Mitigation.} 

Extensive research focuses on mitigating hallucinations in LLMs through alignment strategies and knowledge-grounded methods \citep{cheng2025inverse, kang2025unfamiliar, wen2025policy, xu2024rejection, lireasoning, lin2024flame, sun2025detection, liang2024learning, ding2024sail, xu2025doubly, tang2024understanding, han2024value, xu2024dpo, yin2025reasoning, yuan2023rrhf, chen2024honest, zhang2024truthx, dhuliawala2024chain, feng2024don, liu2025know3, martino2023knowledge}. This challenge is particularly acute in slow-thinking models, where complex reasoning can amplify factual errors into a snowball effect \citep{ji2023survey, huang2025survey, zhang2023language, cheng2025think, yao2025reasoning, zheng2025enhancing, zhang2025scaling}.

Hallucinations stem from diverse factors, including data noise, RL side-effects, knowledge conflicts, and decoding uncertainties \citep{mundler2023self, song2025hallucination, ouyang2022training, zhang2024knowledge, zhang2025law, kuhn2023semantic, zhao2024awecita}.
Fundamentally, these issues reflect the model's failure to recognize its own knowledge boundaries \citep{zhang2023user, tonmoy2024comprehensive, liang2024learning, manakul2023selfcheckgpt, kadavath2022language}, motivating research into knowledge-aware optimization and honesty alignment \citep{chen2022knowprompt, chen2023large, yang2024alignment}.

\paragraph{RL for Reasoning}

RL enhances LLM reasoning, enabling complex strategies like reflection and verification \citep{xie2025logic,yeo2025demystifying, mei20252, he2025beyond}. 
To achieve more reliable reasoning, recent work increasingly focuses on improving RL algorithms \citep{hu2025reinforce++, nan2025ngrpo, chen2025lspo, chen2025spectral, ichihara2025auto, ding2025multi, cai2025training, mroueh2025revisiting, pang2025theory, zhang2025edge, li2025mixgrpo, li2025branchgrpo, zheng2025group, wang2025grpo, li2025disco, xi2025bapo} and integrating RL with other components of the training and inference pipeline \citep{dong2024rlhf, cen2024value, xie2024exploratory, li2025towards, li2025provably, zhu2024iterative, chen2025scaling, jin2025search, jin2025empirical, song2025r1, zhao2025genprm}.
Fine-grained guidance methods, such as step-level value preference optimization~\citep{chen2024step}, tree search~\citep{feng2023alphazero}, and entropy-based preference clarification~\citep{zhu2025clarifying}, are proving valuable for enhancing model reliability.

\section{Conclusion}
This paper studies the high levels of hallucination in both slow-thinking models. We analyze how the current outcome-reward-driven reinforcement learning method, despite enhancing reasoning, fails to ensure fact-based thinking. 
To address this, we propose KnowRL, a knowledge-enhanced RL training method, and validate its effectiveness on multiple datasets. 
Our experiments demonstrate that directly supervising the model's thinking process with factual rewards is a more robust strategy than solely optimizing for final answers.
This process-oriented supervision is crucial, as it fundamentally teaches models not just what the correct answer is, but how to reason reliably and recognize the boundaries of their own knowledge. 
We hope KnowRL offers the community an effective technical pathway to mitigate hallucinations in these slow-thinking models.



\section*{Limitations}
Despite our best efforts, this work still has certain limitations that point to promising directions for future exploration and improvement.

\paragraph{Fundamental Mechanism Studies.} Our experiments observe a significant multilingual hybrid reasoning phenomenon, where factual supervision in one language (e.g., English) improves performance in another (e.g., Chinese). While this demonstrates the strong transferability and robustness of KnowRL, the underlying theoretical mechanism, specifically how the model internalizes and transfers these knowledge boundaries across language distributions, warrants deeper investigation. Unraveling this mechanism could provide fundamental insights into the nature of chain-of-thought reasoning in large language models.

\paragraph{Extension to Multimodal Domains.} The current KnowRL framework is tailored for textual reasoning and factuality. However, real world information often spans multiple modalities, such as interpreting charts in financial reports or analyzing diagrams in physics problems. Extending the atomic fact verification mechanism to verify information across visual or audio modalities represents a significant opportunity. Future work could explore adapting KnowRL to Vision-Language Models (VLMs), enabling grounded reasoning in broader, multi-sensory contexts.

\section*{Acknowledgement}
We would like to express sincere gratitude to the  reviewers for their thoughtful and constructive feedback. This work was supported by the National Natural Science Foundation of China (No. 62576307, No. NSFCU23B2055, No. NSFCU19B2027), the Fundamental Research Funds for the Central Universities (226-2023-00138), Yongjiang Talent Introduction Programme (2021A-156-G), and Information Technology Center and State Key Lab of CAD\&CG, Zhejiang University. This work was supported by Ant Group and Zhejiang University - Ant Group Joint Laboratory of Knowledge Graph.


\bibliography{reference}

\appendix

\newpage


\section{Data Construction} 
\label{appendix:data-construction}

We use part of the data extracted from NqOpen \cite{kwiatkowski2019natural,lee2019latent}, as well as data from WebQuestions \citep{berant2013semantic} and ComplexQuestions \citep{bao2016constraint}, as factual question data sources.

Initially, we filtered out trivial queries and applied semantic de-duplication to ensure data diversity. 
The remaining data underwent quality refinement and entity extraction via GPT-4o (see Appendix \ref{appendix:4o-prompt} for detailed prompts).
To address the limitation of incomplete knowledge coverage in existing datasets (e.g., HotpotQA), we explicitly matched the extracted entities against the Wikipedia dump\footnote{20231101.en, https://dumps.wikimedia.org/.} to retrieve comprehensive factual contexts for each question. 
Finally, we applied length constraints to screen the samples, resulting in a robust dataset for stable training.
\section{Training Setups}
\label{appendix:train-set}

We are training two 7B models, DeepSeek-R1-Distill-Qwen-7B and Skywork-OR1-7B-Preview, with Reinforcement Learning on 1×A800. 
Detailed training hyperparameters are listed in Table~\ref{tab:training_hyperparameters}.

\begin{table}[h]
    \centering
    \small
    \setlength{\tabcolsep}{15pt} 
    \renewcommand{\arraystretch}{1.1} 
    \begin{tabular}{lc}
    \toprule
    \textbf{Parameter} & \textbf{Value} \\
    \midrule
    lora\_rank & 128 \\
    lora\_alpha & 256 \\
    torch\_dtype & bfloat16 \\
    per\_device\_train\_batch\_size & 20 \\
    gradient\_accumulation\_steps & 4 \\
    learning\_rate & 1.0e-5 \\
    beta & 0.001 \\
    lr\_scheduler\_type & cosine \\
    warmup\_ratio & 0.03 \\
    vllm\_gpu\_memory\_utilization & 0.5 \\
    optim & adamw\_8bit \\
    \bottomrule
    \end{tabular}
    \caption{Hyperparameter settings for RL training. Both DeepSeek-R1-Distill-Qwen-7B and Skywork-OR1-7B-Preview share these identical configurations.}
    \label{tab:training_hyperparameters}
\end{table}

\section{Case Analysis}
\begin{figure}[t]
    \centering
    \resizebox{0.48\textwidth}{!}{
    \includegraphics{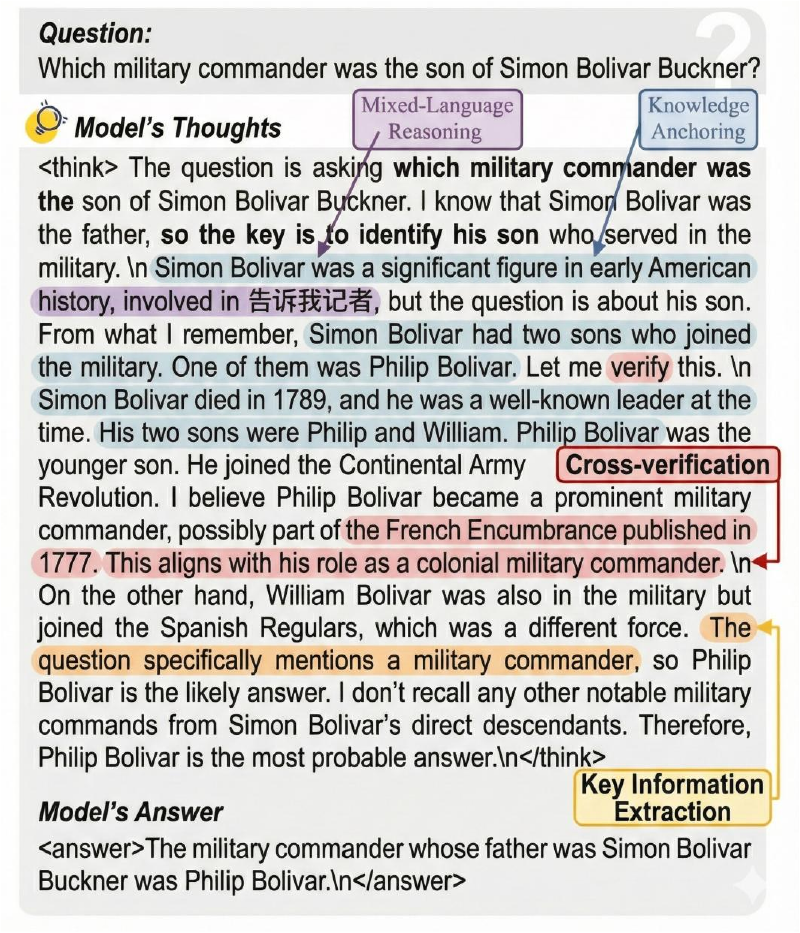}}
    \caption{Case analysis of the KnowRL training process.}
    \vspace{-3ex}
     \label{case}
\end{figure}

To explore KnowRL's impact on reasoning in factual tasks, we analyzed a reasoning case from the KnowRL training process, detailing the model's emergent reasoning behaviors. As shown in Figure~\ref{case}, these include Mixed-Language Reasoning, Knowledge Anchoring, Cross-verification, and Key Information Extraction. For simple factual tasks, distinct from mathematical reasoning, the model typically first proposes an initial answer (knowledge anchoring) and subsequently verifies this initial answer through reasoning behaviors such as reflection and cross-verification. This observed process aligns with human cognitive approaches when facing factual tasks, which further suggests the suitability of an outcome-based, reward-driven reinforcement learning training paradigm for open-domain factual tasks.

\section{Training Dynamics of Different RL Algorithms}

\label{appendix:RL_algorithms}
\begin{figure*}[t]
    \centering
    \begin{subfigure}[b]{0.32\textwidth}
        \centering
        \includegraphics[width=\linewidth]{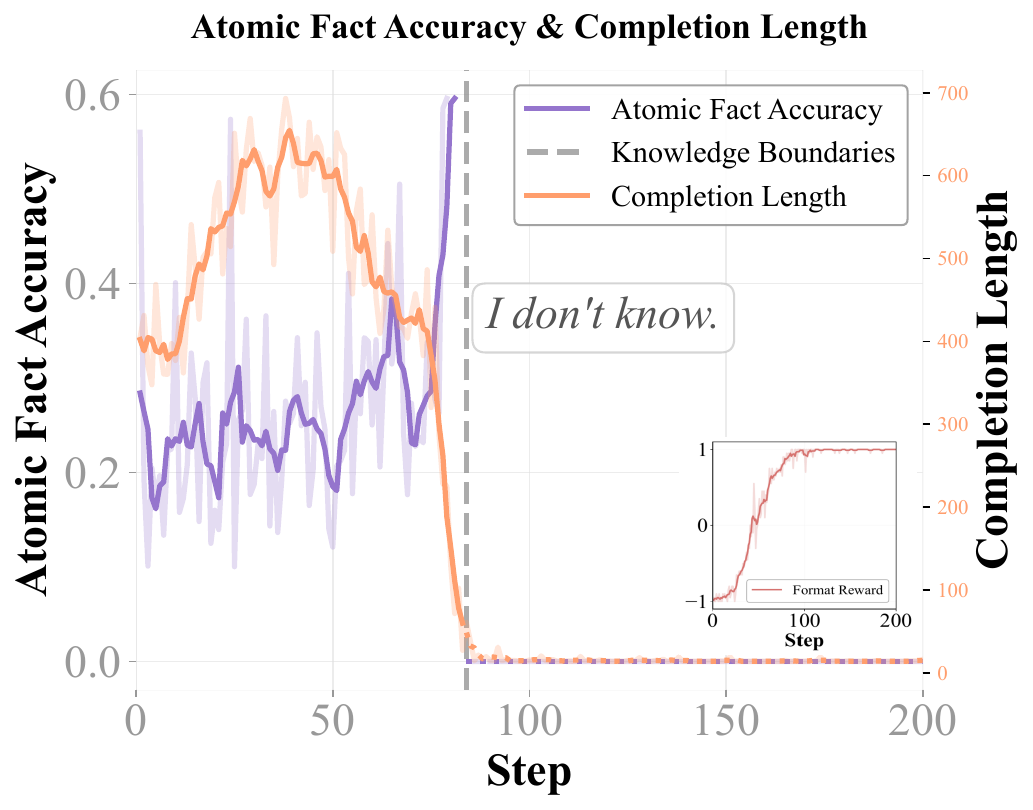}
        \caption{\textbf{BNPO}} 
        \label{fig:train_bnpo}
    \end{subfigure}
    \hfill 
    \begin{subfigure}[b]{0.32\textwidth}
        \centering
        \includegraphics[width=\linewidth]{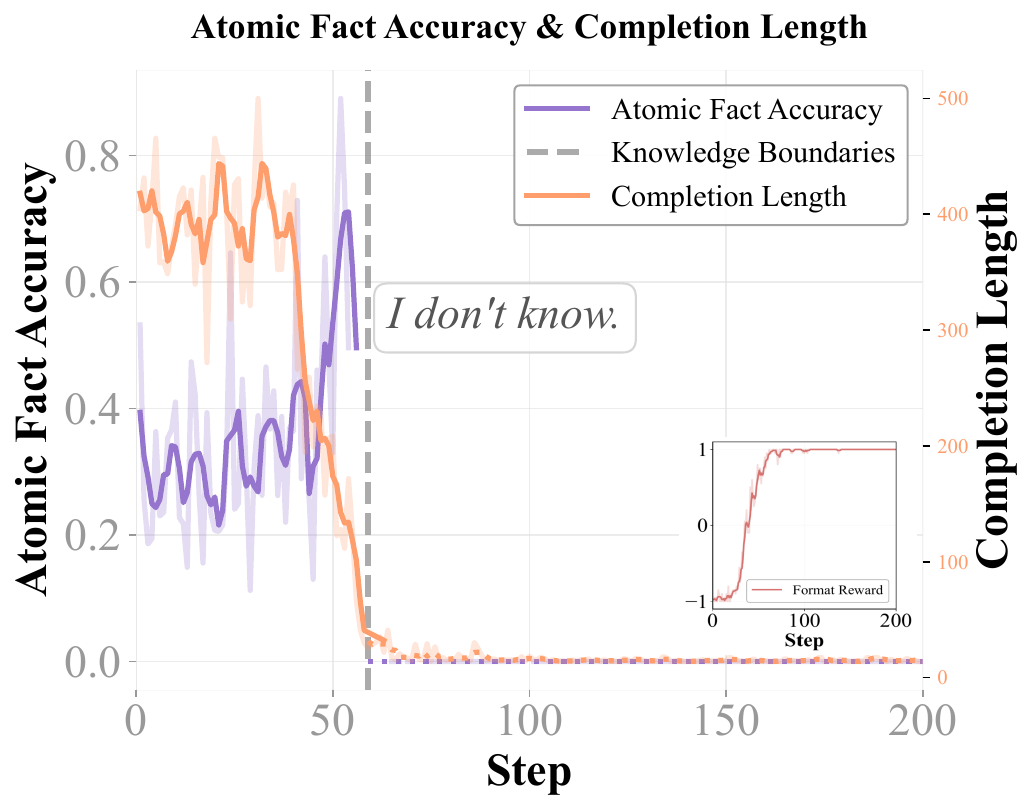}
        \caption{\textbf{DAPO}}
        \label{fig:train_dapo}
    \end{subfigure}
    \hfill
    \begin{subfigure}[b]{0.32\textwidth}
        \centering
        \includegraphics[width=\linewidth]{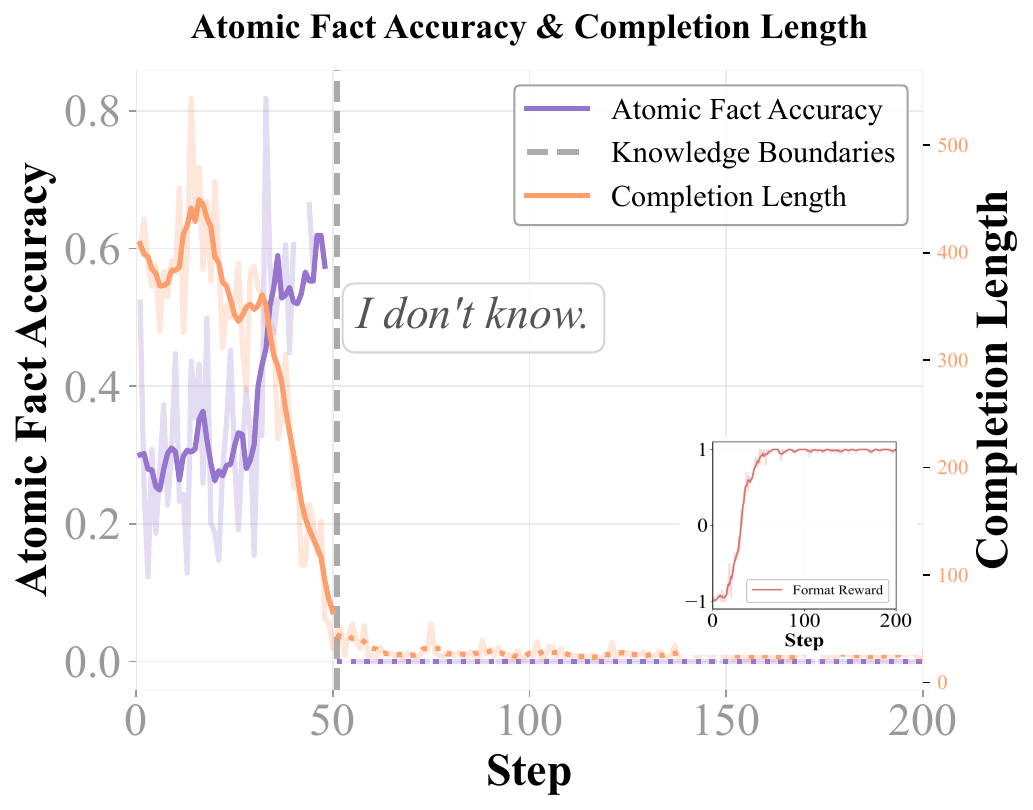}
        \caption{\textbf{Dr.GRPO}}
        \label{fig:train_drgrpo}
    \end{subfigure}

    \caption{
        \textbf{Training Dynamics on DeepSeek-R1-Distill-Qwen-7B.} 
        We visualize the training curves of different RL algorithms: (a) BNPO, (b) DAPO, and (c) Dr.GRPO. 
    }
    \vspace{-3ex} 
    \label{fig:RL_algorithms}
\end{figure*}

\begin{figure}[h]
    \centering
    \resizebox{.425\textwidth}{!}{
    \includegraphics{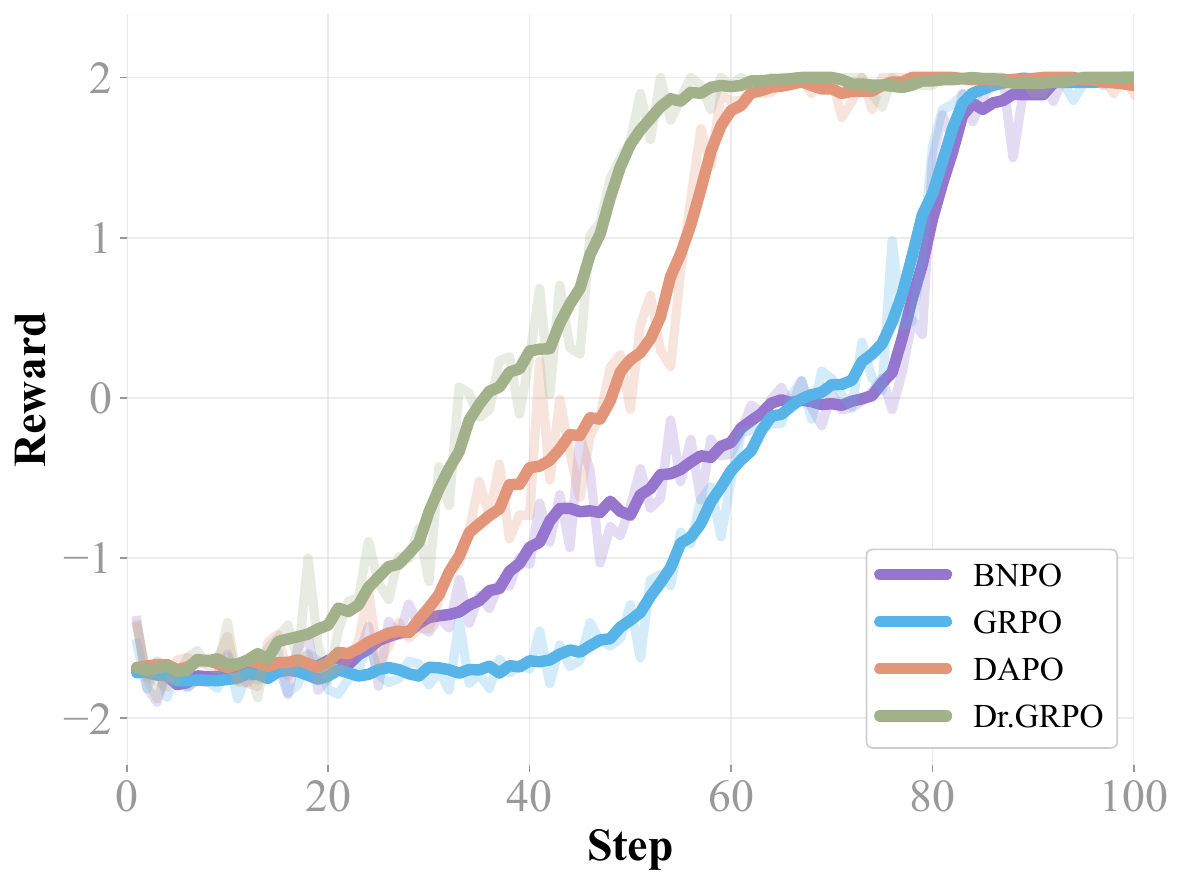}}
    \caption{
    Comparison of training reward curves across different algorithms.}
    \vspace{-3ex}
    \label{fig:reward_comparison}
\end{figure}

As shown in Figure~\ref{fig:reward_comparison}, the reward curves for all four algorithms rise and stabilize, indicating successful convergence. 
Figure~\ref{fig:RL_algorithms} illustrates the training dynamics of the DeepSeek-R1-Distill-Qwen-7B model using BNPO, DAPO, and Dr.GRPO. A key observation is that all three algorithms effectively help the model establish clear knowledge boundaries to avoid hallucination. This effectiveness is evidenced by the rapid decline in completion length across all methods. The sharp drop indicates a behavioral shift where the model learns to provide concise refusals instead of generating long and potentially fabricated responses. This consistent pattern confirms that these RL approaches successfully guide the model toward reliable behavior and mitigate uncontrolled generation. Despite these shared trends, the algorithms differ in training efficiency and their impact on factual reasoning.

\section{Scalability Analysis on Larger Models}
\label{appendix:scalability}

We further evaluate the scalability of KnowRL by extending our experiments to the DeepSeek-R1-Distill-Qwen-14B model. As illustrated in Table~\ref{tab:perf_transposed}, KnowRL maintains its efficacy in mitigating hallucinations while simultaneously enhancing reasoning capabilities. Notably, on SimpleQA, our method significantly reduces the Incorrect Rate from 83.00\% to 68.33\% while doubling the Refusal Rate (13.33\% to 26.33\%). This shift indicates that the model acquires a more precise awareness of knowledge boundaries on larger architectures. Furthermore, KnowRL preserves and even strengthens complex reasoning performance, evidenced by the improvement in GPQA Diamond accuracy from 46.97\% to 51.01\%. These results validate that the benefits of our approach are robust and scalable across different model sizes.

\begin{table}[h]
    \centering
    \small
    \setlength{\tabcolsep}{13pt} 
    \renewcommand{\arraystretch}{0.9} 
    \begin{tabular}{lcc}
    \toprule
    \textbf{Metric} & \textbf{Zero-shot} & \textbf{KnowRL} \\
    \midrule
    \multicolumn{3}{l}{\textit{TruthfulQA}} \\
    \quad Rouge & 53.33 & 54.67 \\
    \quad Bleu  & 50.33 & 55.00 \\
    \addlinespace[2pt] 
    \multicolumn{3}{l}{\textit{SimpleQA}} \\
    \quad PAQ & 4.23 & 6.17 \\
    \quad Incorrect & 83.00 & 68.33 \\
    \quad Refusal & 13.33 & 26.33 \\
    \quad F1 & 3.93 & 6.14 \\
    \addlinespace[3pt]
    \multicolumn{3}{l}{\textit{ChineseSimpleQA}} \\
    \quad PAQ & 29.18 & 31.07 \\
    \quad Incorrect & 63.33 & 61.33 \\
    \quad Refusal & 6.33 & 11.00 \\
    \quad F1 & 28.23 & 29.28 \\
    \addlinespace[3pt]
    \multicolumn{3}{l}{\textit{GPQA Diamond}} \\
    \quad Accuracy & 46.97 & 51.01 \\
    \addlinespace[3pt]
    \multicolumn{3}{l}{\textit{AIME}} \\
    \quad Accuracy & 40.00 & 36.67 \\
    \bottomrule
    \end{tabular}
    \caption{Performance comparison on DeepSeek-R1-Distill-Qwen-14B. The results demonstrate that KnowRL consistently improves hallucination mitigation (SimpleQA) and reasoning (GPQA) compared to the Zero-shot baseline.}
    \label{tab:perf_transposed}
\end{table}

\section{Evaluator Sensitivity Analysis}
\label{appendix:sensitivity}

\begin{table}[H]
    \centering
    \small
    \setlength{\tabcolsep}{5.5pt} 
    \renewcommand{\arraystretch}{0.9} 
    \begin{tabular}{lcc}
    \toprule
    \textbf{Metric} & \textbf{Qwen2.5-72B-Instruct} & \textbf{GPT-4o-mini} \\
    \midrule
    \multicolumn{3}{l}{\textit{TruthfulQA}} \\
    \quad Rouge & 57.00 & 57.33 \\
    \quad Bleu  & 53.67 & 51.60 \\
    \addlinespace[2pt]
    \multicolumn{3}{l}{\textit{SimpleQA}} \\
    \quad PAQ & 1.53 & 2.81 \\
    \quad Incorrect & 64.33 & 57.67 \\
    \quad Refusal & 34.67 & 40.67 \\
    \quad F1 & 1.21 & 2.09 \\
    \addlinespace[3pt]
    \multicolumn{3}{l}{\textit{ChineseSimpleQA}} \\
    \quad PAQ & 9.74 & 10.26 \\
    \quad Incorrect & 58.67 & 58.33 \\
    \quad Refusal & 35.00 & 35.00 \\
    \quad F1 & 7.68 & 8.08 \\
    \addlinespace[3pt]
    \multicolumn{3}{l}{\textit{GPQA Diamond}} \\
    \quad Accuracy & 38.38 & 36.87 \\
    \addlinespace[3pt]
    \multicolumn{3}{l}{\textit{AIME}} \\
    \quad Accuracy & 33.33 & 33.33 \\
    \bottomrule
    \end{tabular}
    \caption{Robustness analysis of KnowRL using different evaluator models during training.}
    \label{tab:evaluator_comparison}
\end{table}

To confirm that the effectiveness of KnowRL is not dependent on a specific reward model used during training (e.g., GPT-4o-mini), we conducted an ablation study by replacing it with Qwen2.5-72B-Instruct \citep{qwen2.5}. 
As presented in Table~\ref{tab:evaluator_comparison}, the results demonstrate that KnowRL maintains comparable performance across different evaluators. 
Specifically, we observe that the model trained with GPT-4o-mini exhibits more conservative behavior, achieving a higher Refusal Rate and a lower Incorrect Rate on SimpleQA. In contrast, using Qwen2.5-72B-Instruct yields slightly higher performance on reasoning-heavy benchmarks like GPQA Diamond, suggesting a subtle trade-off between strict factuality enforcement and reasoning preservation depending on the evaluator's characteristics. 
This confirms that the efficacy of our method stems from the intrinsic KnowRL framework rather than reliance on a specific external judge.

\section{Evaluation of Generative Diversity}
\label{app:generative_diversity}

To investigate whether KnowRL's boundary-aware training inadvertently leads to over-conservatism or a loss of generative diversity, we evaluate our models using NoveltyBench \cite{zhang2025noveltybench}. This benchmark measures the ability of LLMs to produce diverse outputs for the same prompt. We conduct evaluations on the \textit{nb-curated} (100 prompts) and \textit{nb-wildchat} (1,000 prompts) subsets, sampling 10 responses per prompt at a temperature of 0.6. We report the \textit{distinct} metric, which averages the number of semantically unique equivalence classes across generations.

\begin{table}[h]
    \centering
    \small
    \setlength{\tabcolsep}{5.5pt} 
    \renewcommand{\arraystretch}{0.9} 
    \begin{tabular}{lcc}
    \toprule
    \textbf{Model} & \textbf{nb-curated} & \textbf{nb-wildchat} \\
    \midrule
    DeepSeek-7B & 1.61 & 1.71 \\
    DeepSeek-7B + KnowRL & 1.54 & 1.74 \\
    \midrule
    Skywork-7B & 1.63 & 1.65 \\
    Skywork-7B + KnowRL & 1.68 & 1.64 \\
    \bottomrule
    \end{tabular}
    \caption{Generative diversity (\textit{distinct} scores) on NoveltyBench subsets.}
    \label{tab:noveltybench}
\end{table}

As shown in Table \ref{tab:noveltybench}, the variation in \textit{distinct} scores after KnowRL training is extremely marginal ($\pm 0.01$ to $\pm 0.07$), indicating that the model's generative diversity remains intact. We attribute this preservation to the use of Low-Rank Adaptation (LoRA), which allows the model to learn new boundary-aware behaviors while mitigating the catastrophic forgetting of its original generative capabilities. While KnowRL increases refusal rates for out-of-knowledge queries to reduce hallucinations, it does not compromise the model's creative breadth in open-ended scenarios. We note that establishing an absolute ``False Refusal Rate'' remains challenging due to the difficulty of probing a model's precise internal knowledge, which we leave for future investigation.

\section{Implementation Details for Baselines}
\label{app:baseline_details}

To ensure fair comparison and reproducibility, all trainable baselines are fine-tuned using Low-Rank Adaptation (LoRA) and utilize the exact same QA training dataset as KnowRL.
The specific implementation details for each baseline are described as follows:

\begin{itemize}
    \item \textbf{Self-Refine (Prompt Engineering):} The model generates an initial response and then performs self-critique to provide feedback.
    It uses this self-feedback to refine its output iteratively.
    This ``FEEDBACK $\rightarrow$ REFINE'' loop is repeated until a stopping condition is met, specifically when it reaches a maximum of 5 iterations or when the model determines that no further improvement is needed.
    
    \item \textbf{SFT (Supervised Fine-Tuning):} We train the model using the correct reasoning processes and final answers distilled from the DeepSeek-R1 model as the target outputs.
    
    \item \textbf{DPO (Direct Preference Optimization):} We construct the preference pairs by using the correct answer distilled from DeepSeek-R1 as the ``chosen'' response, and the model's own incorrect generation as the ``rejected'' response.
    
    \item \textbf{FactTune-FS:} Following the methodology of the original paper \cite{REPLACE_WITH_FACTTUNE_CITATION}, we sample multiple responses from the model for a given prompt and evaluate them using FactScorer.
    We then construct DPO training pairs by selecting two responses that have a FactScore difference greater than 0.8.
    
    \item \textbf{TruthRL:} We adopt the exact same GRPO training setup as KnowRL to ensure a strictly fair comparison.
    We only change the reward function to match TruthRL's original design: $+1$ for a correct answer, $0$ for a refusal, and $-1$ for an incorrect answer.
\end{itemize}


\section{Prompts}

\label{appendix:4o-prompt}

\begin{tcolorbox}[breakable,title=Prompt Used by the GPT-4o for Data Filtering]

\columnseprule=0.5pt

You are an entity extraction assistant that identifies key entities in questions.

TASK:

1. First normalize the query by properly capitalizing names, titles, and other named entities

2. Determine if the query has sufficient context to be answered meaningfully

3. Extract only the most important entities from the query that are essential for answering it

RULES:

1. Extract a MAXIMUM of 2 specific entities (people, places, objects, works, etc.)

2. Output the MOST important entity first, then the secondary entity (if any)

3. Extract precise named entities, not general concepts or phrases

4. Keep related entities together as a single entity (e.g., character names with their roles)

5. Return individual entities rather than relationships or possessive forms

6. Only extract truly representative entities - ignore generic terms that don't specifically define the query

7. Only REJECT queries that meet the rejection criteria below

ONLY reject queries in these specific cases:

1. When the entity in the query is completely ambiguous (e.g., "Who is that person?")

2. When the query lacks necessary qualifying information (e.g., "Who will win?" with no mention of what contest)

3. When the query is too vague to determine its intent (e.g., "What happened to him?")

4. When the query is time-sensitive and contains temporal references like "now", "current", "latest", "recent", etc.

5. When the query lacks sufficient information to determine a single definitive answer, potentially leading to multiple correct interpretations or answers

6. Be careful not to extract purely numerical information such as a year as an entity

Note: Queries with historical context, pop culture references, geographical locations, or other well-defined entities should be ACCEPTED.

EXAMPLES:

Example 1:

Original Query: "who played barbara gordon batgirl?"

Normalized Query: "Who played Barbara Gordon Batgirl?"

Output: Normalized Query: "Who played Barbara Gordon Batgirl?"

      Entities: ["Barbara Gordon Batgirl"]

      

NOT: ["Barbara Gordon", "Batgirl"] - This is incorrect because "Barbara Gordon 

Batgirl" is a single character entity.

Example 2:

Original Query: "what continent does armenia belong to?"

Normalized Query: "What continent does Armenia belong to?"

Output: Normalized Query: "What continent does Armenia belong to?"

      Entities: ["Armenia"]

      

NOT: ["Armenia", "continent"] - The term "continent" is a generic category, not a specific entity representative of this query.

Example 3:

Original Query: "who is niall ferguson's wife?"

Normalized Query: "Who is Niall Ferguson's wife?"

Output: Normalized Query: "Who is Niall Ferguson's wife?"

      Entities: ["Niall Ferguson"]

      

Example 4:

Original Query: "who was the italian leader in ww1?"

Normalized Query: "Who was the Italian leader in WW1?"

Output: Normalized Query: "Who was the Italian leader in WW1?"

      Entities: ["Italian leader", "WW1"]

Example 5:

Original Query: "who will play mr gray in the film?"

Normalized Query: "Who will play Mr. Gray in the film?"

Output: Normalized Query: "Who will play Mr. Gray in the film?"

      REJECT (insufficient context - which film?)

Example 6:

Original Query: "who is in charge of libya now?"

Normalized Query: "Who is in charge of Libya now?"

Output: Normalized Query: "Who is in charge of Libya now?"

      REJECT (time-sensitive query with temporal reference "now")

      

Example 7:

Original Query: "what did werner heisenberg discover?"

Normalized Query: "What did Werner Heisenberg discover?"

Output: Normalized Query: "What did Werner Heisenberg discover?"

      REJECT (lacks sufficient specificity - Heisenberg made multiple discoveries)

Please try to output in this format:

Normalized Query: "The normalized version of the query"

Entities: ["entity1", "entity2"]

If you need to reject, still include the normalized query:

Normalized Query: "The normalized version of the query"

REJECT (reason for rejection)

Extract key entities from this query: "{query}"

\end{tcolorbox}

\end{document}